\documentclass[sigconf, nonacm]{acmart}
\AtBeginDocument{%
  }

\setcopyright{acmlicensed}
\copyrightyear{2027}
\acmYear{2027}
\acmDOI{10.1145/XXXXXXX.XXXXXXX}
\acmISBN{978-X-XXXX-XXXX-X/2027/05}
\acmConference[WWW 2027]{The ACM Web Conference 2027}{May 10--14, 2027}{Dublin, Ireland}
\acmBooktitle{Proceedings of the ACM Web Conference 2027 (WWW 2027), May 10--14, 2027, Dublin, Ireland}

\usepackage{amsmath}
\usepackage{booktabs}
\usepackage{tikz}
\usetikzlibrary{arrows.meta,positioning}
\begin{document}

\title{NeoResearch: Literature- and Experience-Guided Autonomous Time Series Model Discovery}

\author{Ze Guo}
\affiliation{%
  \institution{State Key Laboratory of Cognitive Intelligence, University of Science and Technology of China}
  \city{Hefei, Anhui Province}
  \country{China}
}
\email{gz1504921411@mail.ustc.edu.cn}

\author{Mingyue Cheng}
\affiliation{%
  \institution{State Key Laboratory of Cognitive Intelligence, University of Science and Technology of China}
  \city{Hefei, Anhui Province}
  \country{China}
}
\email{mycheng@ustc.edu.cn}

\author{Xiaoyu Tao}
\affiliation{%
  \institution{State Key Laboratory of Cognitive Intelligence, University of Science and Technology of China}
  \city{Hefei, Anhui Province}
  \country{China}
}
\email{txytiny@mail.ustc.edu.cn}

\author{Shuo Yu}
\affiliation{%
  \institution{State Key Laboratory of Cognitive Intelligence, University of Science and Technology of China}
  \city{Hefei, Anhui Province}
  \country{China}
}
\email{yu12345@mail.ustc.edu.cn}

\author{Shangtong Ye}
\affiliation{%
  \institution{State Key Laboratory of Cognitive Intelligence, University of Science and Technology of China}
  \city{Hefei, Anhui Province}
  \country{China}
}
\email{yst18349635@mail.ustc.edu.cn}

\begin{abstract}
Developing effective forecasting models for diverse time-series domains requires substantial expert knowledge, evidence from scientific literature on the web, and iterative experimentation. 
However, translating this knowledge and research experience into model designs requires preserving source context and testing applicability under task-specific assumptions.
In this paper, we propose NeoResearch, a literature- and experience-guided autonomous research framework that leverages large language models (LLMs) as researchers to discover lightweight, domain-specific forecasting models. Specifically, NeoResearch establishes an expert-inspired dual-loop research workflow, coupling hypothesis-driven architecture exploration with iterative training and hyperparameter optimization through experimental feedback. To enable knowledge-grounded discovery, the agent autonomously mines scientific literature from the Web, extracting modeling mechanisms, empirical evidence, and applicability conditions to formulate task-grounded, testable architecture hypotheses. Furthermore, a hierarchical experience memory links literature findings to experimental outcomes and consolidates conditional lessons, guiding architecture discovery and training optimization while supporting reuse across forecasting tasks.
Experiments on ten datasets show competitive performance and support dual loops, literature extraction, and memory reuse. Our code is available\footnote{\url{https://github.com/Kawai1Angel/NeoResearch}}.

\end{abstract}
\maketitle

\section{Introduction}
\label{sec:introduction}

\begin{figure}[t]
    \centering
    \includegraphics[width=\linewidth]{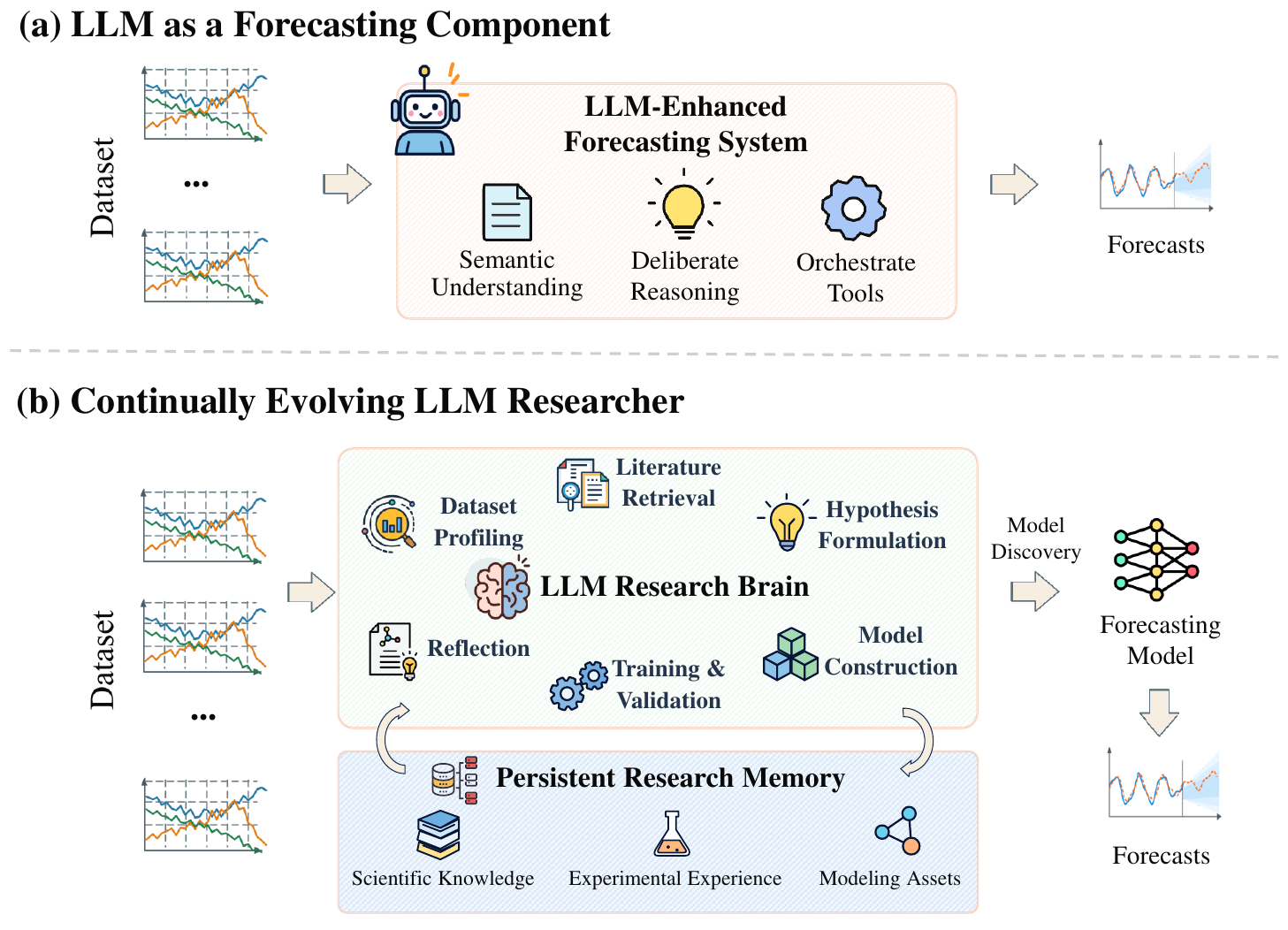}
    \caption{A shared LLM researcher connects Web scientific knowledge with experiments to discover domain-specific forecasters and retain reusable research resources.}
    \Description{The figure contrasts an LLM-enhanced forecasting system with an LLM researcher that connects dataset profiling, literature retrieval, hypotheses, model construction, experiments, and reflection. Persistent memory retains scientific knowledge, experimental experience, and modeling assets.}
    \label{fig:motivation}
\end{figure}

Developing time series forecasting (TSF) models requires matching assumptions to domain-specific temporal dynamics and variable relationships~\cite{zeng2023dlinear,nie2023patchtst,liu2024itransformer}. These choices span preprocessing, temporal representation, and variable interaction, requiring evidence that connects scientific findings to task data. Scientific literature on the web offers mechanisms and empirical findings scattered across papers. Using this content for model design requires preserving source context and checking applicability against target data. Recent studies adapt large language models (LLMs) for TSF through numerical serialization and representation adaptation~\cite{gruver2023llmtime,jin2024timellm}. Further work incorporates reasoning~\cite{zhou2026timer1}, specialized tools~\cite{ye2026tsreasoner}, ensemble optimization~\cite{cao2025conversational}, and accumulated experience~\cite{tao2026memcast} to guide forecasting. These advances motivate using online scientific knowledge and research experience for autonomous model discovery.

Automated forecasting methods search model configurations and architectural operations~\cite{wang2024autots,zhang2023autostl}, while LLM agents extend this automation to model selection, design, and training~\cite{zhao2025timeseriesscientist,trirat2025monaq,shen2026metacaster}. Autonomous research systems connect ideation with experimentation~\cite{tang2025airesearcher,yang2025rdagent}, using literature-derived insights~\cite{wang2024scimon,yang2025nader} and retained experience~\cite{li2025rdagentquant,lyu2026evoscientist}. For domain-specific forecasting, using such resources requires turning scientific content acquired from the web into traceable design decisions whose relevance can be tested on the target task. A mechanism effective in an earlier study may fail under different data conditions. Experimental outcomes must therefore be linked back to the originating findings and retained with conditions for reuse. The central challenge is to translate scientific evidence from the web into testable modeling hypotheses and consolidate experimental outcomes into reusable experience, while preserving source provenance and assessing task-specific applicability.

Addressing this question poses three interconnected challenges. First, architecture exploration must be coordinated with training optimization: poor performance may reflect unsuitable modeling assumptions or inadequate training, requiring evidence to decide whether to refine the current model or reconsider its architecture. Second, online publications report mechanisms and empirical claims in different contexts. Extracted knowledge must preserve source evidence and applicability conditions while yielding testable hypotheses for the target data. Third, experimental histories must be distilled into reusable experience. Useful lessons identify what changed, under which conditions, and with what outcome, while preserving uncertain explanations and allowing reassessment when transferred to new tasks.

We propose NeoResearch, a literature- and experience-guided autonomous research framework in which an LLM discovers lightweight, domain-specific forecasting models that predict independently of the LLM. During preparation, the agent mines scientific literature from the web, recording mechanisms, source passages, empirical evidence, and applicability conditions to ground testable architectural hypotheses. Each investigation yields a forecasting model and reusable experience, as illustrated in Figure~\ref{fig:motivation}. An expert-inspired hierarchical dual-loop workflow connects these hypotheses to experiments. The outer loop proposes and revises architectures using task evidence, literature, and prior experience; the inner loop iteratively optimizes training configurations and hyperparameters and returns validation results and convergence diagnostics. Experimental feedback guides further optimization or architectural reassessment, while structured recipes and persistent state preserve links between source evidence, hypotheses, and experimental outcomes. Hierarchical experience memory consolidates conditional lessons for both loops and subsequent tasks, using new evidence to reassess applicability. Successful interventions guide subsequent designs, while unsuccessful or inconclusive trials are retained with diagnostic context to discourage repeated ineffective changes in later research iterations. Reused architectures are adapted, trained from scratch, and validated on the target task.

Experiments on ten datasets spanning electricity markets, energy systems, and hydrology assess the discovered models and the research process. With human-designed reference-model initialization, NeoResearch outperforms representative forecasting methods and automated model-development baselines across most tasks. Ablations support both research loops and historical experience at both memory levels. Structured literature extraction improves model quality over direct source-text access and excluding literature. Cross-task reuse benefits from retained knowledge, experience, and model designs; structured memory outperforms raw records as source tasks accumulate, although additional sources do not always improve performance. Finally, the MOPEX trajectory illustrates how experiments revise literature- and experience-informed hypotheses, overturning normalization's expected benefit while preserving uncertainty about channel mixing.

Our main contributions are as follows:
\vspace{-0.1cm}
\begin{itemize}
\item We develop a hypothesis-driven dual-loop framework coordinating architecture exploration and training optimization through experimental feedback.
\item We introduce autonomous literature mining that translates mechanisms, source evidence, and applicability conditions into task-specific, testable architectural hypotheses.
\item We introduce hierarchical experience memory consolidating conditional experimental lessons for reuse and revision in research loops and across forecasting tasks.
\end{itemize}
\section{Related Work}
\label{sec:related-work}

\subsection{Time Series Forecasting}
Deep forecasting models encode different assumptions about temporal structure and variable relationships. Recurrent networks capture sequential dynamics through hidden states updated over the observed sequence~\cite{elman1990rnn,hochreiter1997lstm,salinas2020deepar}, while graph networks learn inter-variable dependencies~\cite{wu2020mtgnn} and state-space models represent evolving temporal states~\cite{gu2024mamba,wang2025smamba}. Decomposition-based approaches range from DLinear's linear prediction to TimeMixer's multiscale mixing~\cite{zeng2023dlinear,wang2024timemixer}. TimesNet and ConvTimeNet use convolutional designs to capture temporal variation~\cite{wu2023timesnet,cheng2025convtimenet}. Transformer architectures emphasize temporal patches, variable tokens, and exogenous information in PatchTST, iTransformer, and TimeXer, respectively~\cite{nie2023patchtst,liu2024itransformer,wang2024timexer}. Their suitability depends on task characteristics, including the temporal patterns, variable relationships, and available covariates that each architecture is designed to exploit. Automated methods search forecasting architectures and configurations~\cite{wang2024autots,zhang2023autostl}, and LLM agents extend automation to model selection, design, and training. TimeSeriesScientist combines diagnostics, model selection, and validation-guided configuration~\cite{zhao2025timeseriesscientist}; MONAQ generates resource-constrained architectures using LLM estimates for search-time evaluation~\cite{trirat2025monaq}; and MetaCaster combines domain-aware synthetic data with lightweight-model training and harness optimization~\cite{shen2026metacaster}. NeoResearch follows this model-discovery direction, coordinating architecture hypotheses with iterative training optimization to produce lightweight, domain-specific forecasters that predict independently of the LLM.

\begin{figure*}[t]
    \centering
    \includegraphics[width=\linewidth]{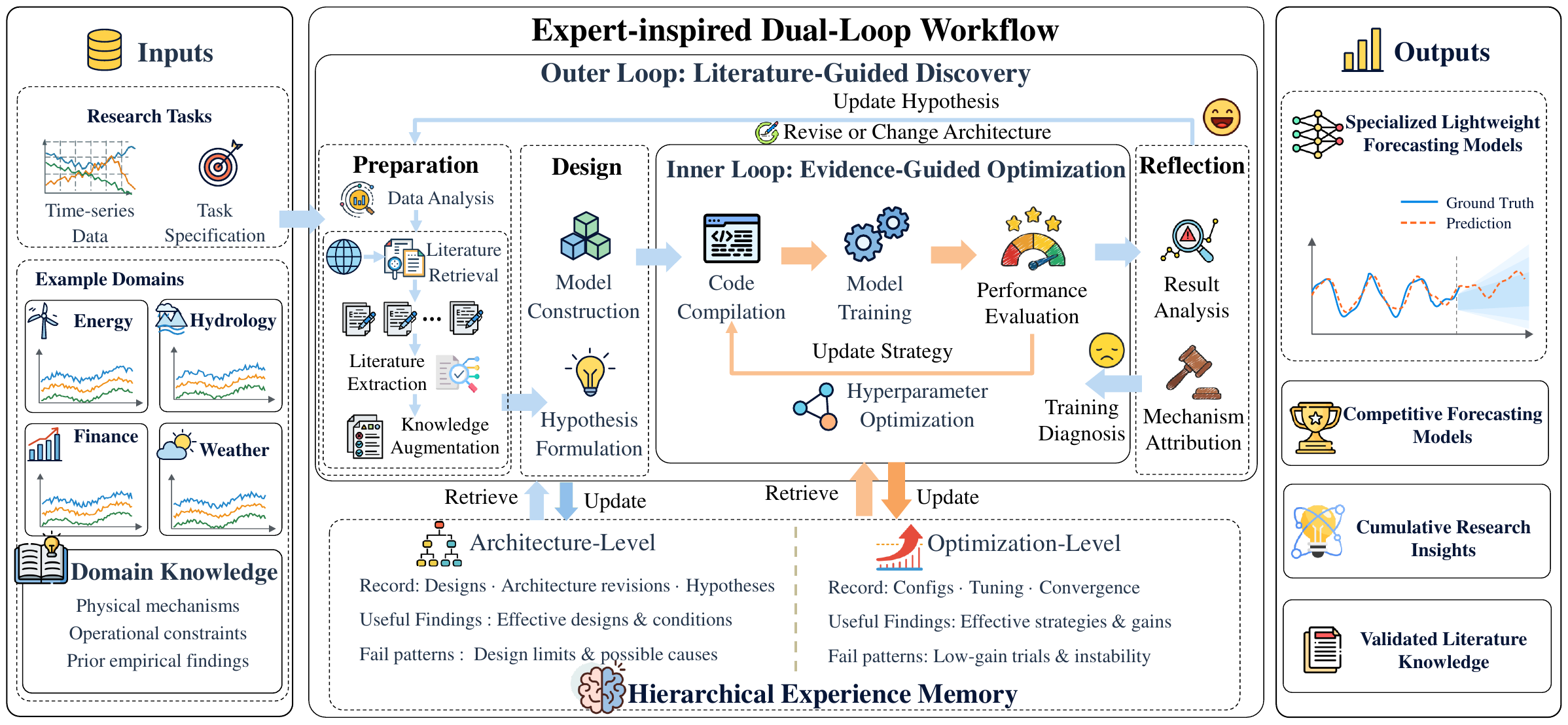}
    \caption{NeoResearch couples literature-guided architecture discovery with evidence-guided training optimization. Hierarchical experience memory guides both loops within and across forecasting tasks.}
    \Description{An LLM researcher profiles data, retrieves literature, formulates hypotheses, constructs models, trains and validates them, and reflects on evidence. Research tools, model recipes, persistent state, and accumulated memory support the cycle.}
    \label{fig:framework}
\end{figure*}

\subsection{LLM-Based Autonomous Research}

LLM-based research connects literature-grounded ideation, experimentation, and experience reuse. For ideation, SciMON retrieves inspirations and revises proposals by comparing them with prior work for novelty~\cite{wang2024scimon}. ResearchAgent uses academic-graph retrieval, mined scientific concepts, and reviewing agents to refine problems, methods, and experiment designs~\cite{baek2025researchagent}.
Integrated systems connect proposals to implementation and evaluation. CodeScientist combines curated articles and codeblocks to build and run experiments, with human selection and feedback~\cite{jansen2025codescientist}. AI-Researcher connects papers and code repositories to algorithm implementation and iterative validation~\cite{tang2025airesearcher}. NADER organizes paper-informed architecture modifications in a search tree, trains candidate networks, and retains successful and failed design experience~\cite{yang2025nader}. The AI Scientist connects ideation, experiments, and reporting~\cite{lu2024aiscientist}, while AIDE explores machine-learning solutions through code-space tree search~\cite{jiang2025aide}. R\&D-Agent separates hypothesis exploration from solution development, using experiment memory and execution feedback to refine candidates~\cite{yang2025rdagent}.
Experience memory supports subsequent investigations. R\&D-Agent-Quant retains hypothesis--feedback histories and task--code--feedback records for subsequent iterations~\cite{li2025rdagentquant}, while EvoScientist maintains research and experimentation memories for reuse under new research goals~\cite{lyu2026evoscientist}. These approaches retain different forms of evidence to support hypothesis selection, implementation, and experimental strategy. NeoResearch focuses on literature- and experience-guided discovery of domain-specific forecasting architectures.

\section{The Proposed NeoResearch}
\label{sec:methods}

\subsection{Problem Formulation}
\label{sec:research-setting}

For a domain-specific forecasting task $T$, let $X\in\mathbb{R}^{L\times C}$ denote $L$ historical observations of $C$ variables and $Y\in\mathbb{R}^{H}$ the target sequence over a forecast horizon $H$. The rows of $X$ index historical time steps and the columns index input variables. Each element of $Y$ is the target value at one forecast step. A candidate forecaster $f$ maps $X$ to predictions of $Y$. Guided by scientific literature and accumulated research experience, NeoResearch explores model architectures and training configurations under a fixed research budget and selects $\hat{f}=\operatorname{arg\,min}_{f\in\mathcal{F}}\mathcal{L}_{\mathrm{val}}(f)$, where $\mathcal{F}$ contains all candidate models successfully trained and evaluated on validation data within that budget, $\hat{f}$ is the final selected model, and $\mathcal{L}_{\mathrm{val}}$ denotes validation MSE under the task's evaluation protocol. Research findings are retained with their evidence  to support subsequent tasks. We consider single-target forecasting with historical covariates only; test data are excluded from research decisions.

\subsection{Framework Overview}
\label{sec:framework-overview}
\label{sec:research-cycle}

NeoResearch employs a general-purpose LLM as an autonomous researcher that translates scientific literature and experimental experience into task-specific forecasting models. As illustrated in Figure~\ref{fig:framework}, an expert-inspired hierarchical dual-loop workflow connects knowledge-grounded hypothesis formulation with experimental evaluation. The outer architecture discovery loop mines literature, interprets task evidence, and retrieves relevant experience to propose architectural hypotheses. The inner training optimization loop optimizes each candidate's training configuration and provides validation results and diagnostics for hypothesis assessment. Feedback guides architectural refinement or alternative hypotheses, while reflection consolidates experimental outcomes into conditional research experience. This memory supports decisions within the current task and transfers applicable findings to subsequent tasks. Persistent research state links hypotheses, designs, experiments, and assessments throughout the investigation. Expert inspiration refers to this research organization; continuous human guidance is a separate interaction mode. The resulting lightweight forecasting models perform numerical prediction independently of the LLM.

\subsection{Outer Loop: Literature-Guided Discovery}
\label{sec:architecture-discovery}

The outer architecture discovery loop connects task characteristics to scientific knowledge and prior research experience. Data diagnostics characterize missing values, distributions, temporal dynamics, and relationships among variables. Guided by these observations and domain information, the researcher autonomously retrieves relevant scientific literature from the web and extracts modeling mechanisms, empirical evidence, and applicability conditions. Reported findings are interpreted in their original experimental context, and their assumptions are checked against the target data and forecasting objective. Retrieved experience provides additional evidence about successful designs, failed interventions, and their conditions. Together, these sources motivate task-specific, testable architectural hypotheses: temporal dependence may support a temporal backbone, while variable relationships may motivate an interaction module. Source references, supporting observations, and explicit applicability assumptions are retained so that subsequent experiments can reassess the design rationale. Unresolved results can prompt additional diagnostics or literature retrieval.

Each hypothesis specifies a proposed modeling mechanism, its expected effect, and observations that would challenge the explanation. Each hypothesis records its applicability conditions, allowing negative results to be examined for violated assumptions. At outer iteration $i$, the researcher translates a selected hypothesis into a candidate design $d_i$ comprising preprocessing and architecture. Design choices include the modeling paradigm, temporal backbone, variable interaction, and module composition. The inner loop then evaluates the candidate under iteratively refined training configurations. Outer-loop reflection analyzes validation results and experimental history to assess the proposed mechanism, update the hypothesis, and refine or replace the architecture. Controlled comparisons identify a reference model and isolate declared changes while holding other architecture and training settings fixed. Architecture discovery therefore proceeds through repeated interaction between literature-grounded reasoning and experimental assessment. A structured model specification, referred to as a recipe, records the design and training choices for each experiment:
\begin{equation}
    r_{i,j} = (p_i, a_i, o_{i,j}, c_{i,j}),
    \label{eq:model-recipe}
\end{equation}
where $p_i$ specifies input configuration and preprocessing, $a_i$ defines the architecture, $o_{i,j}$ is the training loss, and $c_{i,j}$ specifies the training configuration, including the optimizer and hyperparameters, for inner-loop attempt $j$. The design $d_i=(p_i,a_i)$, including architectural dimensions and module choices, remains fixed as $j$ varies. Training loss $o_{i,j}$ may vary across attempts, whereas $\mathcal{L}_{\mathrm{val}}$ consistently measures validation MSE for model selection. Model construction checks representation compatibility, branch composition, and consistency with the proposed intervention. Recipes link hypotheses to executable experiments, supporting comparison between intended and observed effects. Saved models retain weights, recipes, preprocessing, inference code, and evaluation metadata.

\subsection{Inner Loop: Evidence-Guided Optimization}
\label{sec:training-optimization}
\label{sec:research-harness}
The inner training optimization loop conducts construction, compatibility, and preliminary execution checks for candidate design $d_i$ before training and validation. With the design and architectural hypothesis fixed, the researcher iteratively adjusts the learning rate, batch size, regularization, scheduling strategy, and training loss using current experimental feedback and relevant training experience. Validation metrics, convergence curves, and error analysis inform these choices and provide evidence for subsequent architectural assessment. Changes to the backbone, module composition, or preprocessing require an outer-loop design revision. Implementation errors are corrected without altering the hypothesis or agreed experimental conditions; incompatible designs return to the outer loop. Each attempt $j$ produces a record $e_{i,j}$ containing the executed recipe, training conditions, measurements, diagnostics, and execution status. The history $E_i=\{e_{i,j}\}_j$ preserves completed and failed attempts. Model selection considers only successfully trained and validated candidates:
\begin{equation}
    \hat{f}_i = \underset{f\in\mathcal{F}_i}{\operatorname{arg\,min}}\;\mathcal{L}_{\mathrm{val}}(f),
    \label{eq:inner-selection}
\end{equation}
where $\mathcal{F}_i\subseteq\mathcal{F}$ contains the models successfully trained and evaluated on validation data for design $d_i$, and $\hat{f}_i$ is the best model selected for that design. The selected model and experiment history inform whether further training optimization is justified or the architectural hypothesis requires reassessment. Optimization remains subject to convergence, validation gains, and the shared iteration, time, and resource budgets. Architectural revisions do not reset consumed resources.

The execution framework maintains persistent task state $S$ linking hypotheses, recipes, runs, model artifacts, and assessments. Experimental records preserve data and training settings, validation metrics, convergence curves, and execution status, enabling comparisons within and across designs. These records support hypothesis assessment and experience consolidation, including the interpretation of unsuccessful interventions. Decision logs preserve the research trajectory, allowing prior failures and unresolved questions to inform later iterations. Conditional lessons remain linked to experimental records of source designs, adaptations, and outcomes. Failed runs remain distinct from completed evaluations. Cached measurements are checked against the recipe, data, training settings, and execution conditions before reuse. Verified cached results do not constitute independent evidence or increment the count of consecutive iterations without improvement. When research resumes, the framework verifies inputs, artifacts, and execution ownership while retaining the experiment history and consumed budget.

\subsection{Hierarchical Experience Memory}
\label{sec:research-memory}

NeoResearch maintains hierarchical experience memory $M$ for architecture- and optimization-level reflection, guiding decisions within and across tasks while mitigating stagnation. Entries link literature findings, model recipes, and experimental outcomes to diagnostic interpretations and applicability conditions. Task-level findings are consolidated as:
\begin{equation}
    M' = G(M, S),
    \label{eq:resource-consolidation}
\end{equation}
where $M'$ is the updated memory, $S$ is the completed task's research state, and $G$ denotes consolidation. Subsequent tasks retrieve applicable findings for target-task validation; contradicted or unknown conditions preclude direct adoption. Original literature records and the LLM's parameters remain unchanged.

\begin{table*}[t]
\centering
\caption{Forecasting performance of the discovered domain-specific models. NeoResearch achieves the lowest error in 19 of 20 dataset--metric comparisons. Bold and underlined values indicate the best and second-best results, respectively.}
\label{tab:main-results}
\label{page:experiments-start}
\setlength{\tabcolsep}{1.5pt}
\renewcommand{\arraystretch}{0.85}
\resizebox{\textwidth}{!}{%
\begin{tabular}{lrrrrrrrrrrrrrrrrrrrr}
\toprule
Method & \multicolumn{2}{c}{ETTh1} & \multicolumn{2}{c}{GEF-L} & \multicolumn{2}{c}{Windy} & \multicolumn{2}{c}{Sunny} & \multicolumn{2}{c}{MOPEX} & \multicolumn{2}{c}{BE} & \multicolumn{2}{c}{DE} & \multicolumn{2}{c}{FR} & \multicolumn{2}{c}{NP} & \multicolumn{2}{c}{PJM} \\
\cmidrule(lr){2-3} \cmidrule(lr){4-5} \cmidrule(lr){6-7} \cmidrule(lr){8-9} \cmidrule(lr){10-11} \cmidrule(lr){12-13} \cmidrule(lr){14-15} \cmidrule(lr){16-17} \cmidrule(lr){18-19} \cmidrule(lr){20-21}
 & MSE & MAE & MSE & MAE & MSE & MAE & MSE & MAE & MSE & MAE & MSE & MAE & MSE & MAE & MSE & MAE & MSE & MAE & MSE & MAE \\
\midrule
ARIMA & 0.1258 & 0.2692 & 2.6140 & 1.1724 & 1.5244 & \underline{0.7946} & 2.1279 & 0.9284 & 2.0232 & 0.8539 & 2.7734 & 0.8377 & 1.4358 & 0.8512 & 11.6593 & 1.2454 & 0.7585 & 0.5777 & 1.0326 & 0.7560 \\
Prophet & 2.7527 & 1.4975 & 2.3506 & 1.0190 & 2.1633 & 1.1161 & 0.3636 & 0.4571 & 1.5570 & 0.8254 & 2.9572 & 0.9921 & 1.2569 & 0.7406 & 8.7114 & 1.5332 & 4.9137 & 1.9780 & 0.5472 & 0.5890 \\
GNN & 0.1710 & 0.3282 & 0.3483 & 0.4194 & 1.3559 & 0.8382 & 0.3204 & 0.3615 & 1.1744 & \underline{0.6094} & 1.7677 & 0.5849 & 0.9764 & 0.6892 & 5.2215 & 0.8461 & 0.4129 & 0.4228 & 0.2301 & 0.3596 \\
MTGNN & 0.1754 & 0.3280 & 0.3421 & 0.4157 & 1.8594 & 1.0386 & 0.3711 & 0.3940 & \underline{1.1581} & 0.6287 & 1.5649 & 0.5537 & 2.0645 & 1.0449 & 6.0711 & 0.8063 & 0.5049 & 0.4874 & 0.2250 & 0.3514 \\
RNN & 0.1339 & 0.2826 & 0.3665 & 0.4306 & 1.9703 & 1.0033 & 0.5035 & 0.4779 & 1.2899 & 0.6809 & 1.7545 & 0.5831 & 0.8215 & 0.6194 & 6.0330 & 0.8054 & 0.4425 & 0.4521 & 0.2538 & 0.3701 \\
LSTM & 0.3040 & 0.4590 & 0.3773 & 0.4454 & 1.5528 & 0.9778 & 0.6512 & 0.5681 & 1.3071 & 0.6815 & 3.6777 & 0.9915 & 0.8508 & 0.6080 & 5.8089 & 0.7482 & 0.5449 & 0.4880 & 0.2411 & 0.3599 \\
DeepAR & 0.1082 & 0.2558 & 0.3627 & 0.4295 & \underline{1.2920} & 0.8204 & 0.7127 & 0.4818 & 1.4699 & 0.6281 & 1.7251 & 0.5263 & \underline{0.8030} & \underline{0.6018} & 6.0681 & 0.8572 & 0.9570 & 0.7178 & 0.2226 & 0.3401 \\
Mamba & 0.1265 & 0.2795 & 0.4522 & 0.4931 & 1.6849 & 0.8810 & 0.3213 & 0.3623 & 1.3003 & 0.6815 & 2.2539 & 0.5820 & 1.2701 & 0.7626 & 6.3837 & 0.8629 & 0.4718 & 0.4305 & 0.2348 & 0.3479 \\
S-Mamba & 0.1077 & 0.2493 & 0.3520 & 0.4254 & 1.4024 & 0.8072 & 0.3141 & 0.3697 & 1.3262 & 0.6898 & 1.5365 & 0.5359 & 0.8214 & 0.6080 & 5.4481 & 0.8019 & 0.3994 & 0.3913 & 0.2185 & 0.3408 \\
DLinear & 0.1049 & 0.2482 & 0.3695 & 0.4445 & \textbf{1.2785} & 0.8107 & \underline{0.3119} & 0.3939 & 1.2572 & 0.6675 & 1.6316 & 0.5625 & 0.9382 & 0.6590 & 6.0839 & 0.9333 & 0.4556 & 0.4261 & 0.2585 & 0.3627 \\
TimeMixer & 0.1055 & 0.2510 & 0.4046 & 0.4619 & 1.3718 & 0.7969 & 0.3222 & \underline{0.3598} & 1.4624 & 0.7241 & 1.4927 & 0.5165 & 0.8099 & 0.6199 & 5.7979 & 0.7887 & \underline{0.3842} & 0.3824 & 0.2130 & 0.3376 \\
ConvTimeNet & 0.1040 & 0.2473 & 0.3760 & 0.4372 & 1.3877 & 0.7955 & 0.3325 & 0.3678 & 1.5215 & 0.8029 & 1.7362 & 0.6143 & 0.8857 & 0.6439 & 5.8776 & 0.8712 & 0.4146 & 0.4044 & 0.2544 & 0.3635 \\
TimesNet & 0.1070 & 0.2540 & 0.3828 & 0.4474 & 1.4368 & 0.8064 & 0.3473 & 0.3818 & 1.4196 & 0.7100 & 1.6829 & 0.5574 & 0.8985 & 0.6514 & 6.1616 & 0.8882 & 0.4706 & 0.4303 & 0.2306 & 0.3454 \\
PatchTST & 0.1060 & 0.2502 & 0.3702 & 0.4365 & 1.3710 & 0.8025 & 0.3307 & 0.3616 & 1.3829 & 0.7067 & 1.4701 & 0.5200 & 0.8162 & 0.6122 & 5.1874 & 0.7330 & 0.3893 & \underline{0.3786} & 0.2211 & 0.3414 \\
iTransformer & \underline{0.1022} & \underline{0.2436} & 0.3565 & 0.4266 & 1.4412 & 0.7986 & 0.3175 & 0.3726 & 1.3043 & 0.6879 & 1.6322 & 0.5575 & 0.8606 & 0.6332 & 6.0580 & 0.8921 & 0.4020 & 0.3909 & 0.2123 & 0.3358 \\
TimeXer & 0.1049 & 0.2441 & 0.3520 & 0.4263 & 1.3679 & 0.7947 & 0.3355 & 0.3773 & 1.2044 & 0.6400 & 1.6470 & 0.5537 & 0.9687 & 0.6793 & \underline{5.1838} & 0.7670 & 0.4329 & 0.4119 & 0.2159 & 0.3331 \\
AIDE & 0.1041 & 0.2481 & \underline{0.3384} & \underline{0.4119} & 1.6653 & 0.9038 & 0.3261 & 0.3841 & 1.1838 & 0.6812 & \underline{1.4386} & \underline{0.4764} & 0.8087 & 0.6063 & 5.4238 & \underline{0.7262} & 0.4076 & 0.4105 & 0.2200 & 0.3366 \\
Optuna-TPE & 0.1035 & 0.2468 & 0.3549 & 0.4293 & 1.4479 & 0.8453 & 0.3433 & 0.3932 & 1.2881 & 0.6615 & 1.5058 & 0.5354 & 0.8228 & 0.6097 & 5.3778 & 0.7977 & 0.3959 & 0.3854 & \underline{0.2116} & \underline{0.3269} \\
\midrule
\textbf{NeoResearch} & \textbf{0.1004} & \textbf{0.2434} & \textbf{0.3324} & \textbf{0.4114} & 1.3406 & \textbf{0.7867} & \textbf{0.3089} & \textbf{0.3552} & \textbf{0.9812} & \textbf{0.5443} & \textbf{1.3807} & \textbf{0.4637} & \textbf{0.7912} & \textbf{0.5948} & \textbf{5.0887} & \textbf{0.6675} & \textbf{0.3546} & \textbf{0.3618} & \textbf{0.1849} & \textbf{0.3093} \\
\bottomrule
\end{tabular}%
}
\end{table*}

\subsubsection{Memory for Architecture Discovery}
The outer loop retrieves mechanisms, prior hypotheses, and architectural outcomes matched to the task. Experience informs which designs to adapt, which assumptions to test, and when limited gains justify exploring an alternative hypothesis. Adapted recipes are trained with newly initialized weights and validated on the target task. Reflection checks implementation, evaluation consistency, training adequacy, and confounding changes before revising a finding's interpretation or scope. Mechanistic attribution requires controlled comparisons; ambiguous or conflicting outcomes remain unresolved.

\subsubsection{Memory for Training Optimization}
The inner loop retrieves training configurations, convergence patterns, and tuning outcomes relevant to the current design. Successful strategies guide configuration choices; unsuccessful paths discourage repeated ineffective adjustments. Validation and convergence evidence informs whether to continue optimization or revisit architectural assumptions. Diagnoses, decisions, and subsequent outcomes are recorded with applicability conditions for reuse within and across tasks. Execution failures and performance plateaus support diagnosis but do not directly refute an architecture.


\section{Experiments}
\label{sec:experiments}

\subsection{Experimental Setup}
\label{sec:experimental-setup}

\subsubsection{Datasets}
We evaluate ten multivariate-input, single-target forecasting tasks spanning electricity markets, energy systems, and hydrology: BE, DE, FR, NP, and PJM from the electricity price forecasting (EPF) benchmark~\cite{lago2021epf}; ETTh1~\cite{zhou2021informer}; GEFCom2014-L (GEF-L)~\cite{hong2016gefcom2014}; Windy and Sunny; and MOPEX~\cite{schaake2006mopex}. EPF tasks use 168 past hourly observations to forecast 24 hourly prices; the remaining tasks use 96 observations to forecast 96 target values. All covariates are restricted to observed-past inputs; known-future and static covariates are unavailable for forecasting. Appendix~\ref{app:datasets} provides dataset descriptions, sampling frequencies, split settings, and detailed statistics for each dataset.

\begin{figure}[t]
    \centering
        \includegraphics[width=\linewidth]{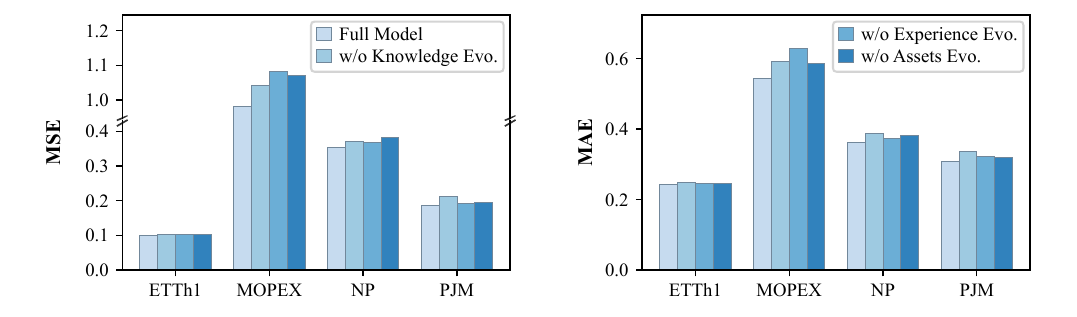}
            \caption{Ablation of cross-task resource reuse. Restricting reuse worsens forecasts.}
        \Description{Full NeoResearch outperforms variants that block stored knowledge cards, exclude cross-task experience, or disable modeling-asset reuse. MOPEX shows the largest gaps.}
        \label{fig:cross-dataset-evolution}
\end{figure}

\begin{table*}[t]
\centering
\caption{Dual-loop ablation. Full NeoResearch achieves lower MSE and MAE than both variants on all ten tasks. Bold and underlined values indicate the best and second-best results, respectively.}
\label{tab:evidence-ablation}
\setlength{\tabcolsep}{1.5pt}
\renewcommand{\arraystretch}{0.85}
\resizebox{\textwidth}{!}{%
\begin{tabular}{lrrrrrrrrrrrrrrrrrrrr}
\toprule
Method & \multicolumn{2}{c}{ETTh1} & \multicolumn{2}{c}{GEF-L} & \multicolumn{2}{c}{Windy} & \multicolumn{2}{c}{Sunny} & \multicolumn{2}{c}{MOPEX} & \multicolumn{2}{c}{BE} & \multicolumn{2}{c}{DE} & \multicolumn{2}{c}{FR} & \multicolumn{2}{c}{NP} & \multicolumn{2}{c}{PJM} \\
\cmidrule(lr){2-3} \cmidrule(lr){4-5} \cmidrule(lr){6-7} \cmidrule(lr){8-9} \cmidrule(lr){10-11} \cmidrule(lr){12-13} \cmidrule(lr){14-15} \cmidrule(lr){16-17} \cmidrule(lr){18-19} \cmidrule(lr){20-21}
 & MSE & MAE & MSE & MAE & MSE & MAE & MSE & MAE & MSE & MAE & MSE & MAE & MSE & MAE & MSE & MAE & MSE & MAE & MSE & MAE \\
\midrule
w/o Outer Loop & \underline{0.1025} & \underline{0.2473} & \underline{0.3427} & 0.4210 & 1.6799 & 0.9156 & \underline{0.4470} & 0.4309 & 1.5570 & 0.7740 & 1.7829 & 0.6017 & 1.3194 & 0.7780 & 6.5556 & 1.0450 & 0.3783 & 0.3766 & 0.2115 & 0.3270 \\
w/o Inner Loop & 0.1051 & 0.2493 & 0.3442 & \underline{0.4209} & \underline{1.4145} & \underline{0.8415} & 0.4783 & \underline{0.4155} & \underline{1.0301} & \underline{0.5655} & \underline{1.4870} & \underline{0.4860} & \underline{0.8276} & \underline{0.6108} & \underline{5.5161} & \underline{0.7002} & \underline{0.3759} & \underline{0.3737} & \underline{0.1883} & \underline{0.3196} \\
\midrule
\textbf{NeoResearch} & \textbf{0.1004} & \textbf{0.2434} & \textbf{0.3324} & \textbf{0.4114} & \textbf{1.3406} & \textbf{0.7867} & \textbf{0.3089} & \textbf{0.3552} & \textbf{0.9812} & \textbf{0.5443} & \textbf{1.3807} & \textbf{0.4637} & \textbf{0.7912} & \textbf{0.5948} & \textbf{5.0887} & \textbf{0.6675} & \textbf{0.3546} & \textbf{0.3618} & \textbf{0.1849} & \textbf{0.3093} \\
\bottomrule
\end{tabular}%
}
\end{table*}

\begin{table}[tb]
    \centering
    \caption{Hierarchical memory ablations. Removing memory for architecture discovery, training optimization, or both increases forecasting errors.}
    \label{tab:memory-ablation}
    \setlength{\tabcolsep}{1.5pt}
    \renewcommand{\arraystretch}{0.85}
    \resizebox{\linewidth}{!}{%
    \begin{tabular}{@{}lrrrrrrrr@{}}
    \toprule
    Method & \multicolumn{2}{c}{ETTh1} & \multicolumn{2}{c}{MOPEX} & \multicolumn{2}{c}{NP} & \multicolumn{2}{c}{PJM} \\
    \cmidrule(lr){2-3} \cmidrule(lr){4-5} \cmidrule(lr){6-7} \cmidrule(lr){8-9}
     & MSE & MAE & MSE & MAE & MSE & MAE & MSE & MAE \\
    \midrule
    w/o Architecture-level & \underline{0.1032} & \underline{0.2464} & \underline{1.0974} & \underline{0.6080} & \underline{0.3894} & \underline{0.3787} & 0.2206 & 0.3333 \\
    w/o Optimization-level & 0.1057 & 0.2507 & 1.2361 & 0.6767 & 0.3933 & 0.3850 & \underline{0.2013} & \underline{0.3217} \\
    w/o Memory & 0.1060 & 0.2491 & 1.3206 & 0.6863 & 0.4092 & 0.3951 & 0.2122 & 0.3338 \\
    \midrule
    \textbf{NeoResearch} & \textbf{0.1004} & \textbf{0.2434} & \textbf{0.9812} & \textbf{0.5443} & \textbf{0.3546} & \textbf{0.3618} & \textbf{0.1849} & \textbf{0.3093} \\
    \bottomrule
    \end{tabular}%
    }
\end{table}

\subsubsection{Baselines}
We compare 16 forecasting baselines: ARIMA~\cite{box2015arima} and Prophet~\cite{taylor2018prophet}; graph-based GNN~\cite{scarselli2009gnn} and MTGNN~\cite{wu2020mtgnn}; recurrent RNN~\cite{elman1990rnn}, LSTM~\cite{hochreiter1997lstm}, and DeepAR~\cite{salinas2020deepar}; state-space Mamba~\cite{gu2024mamba} and S-Mamba~\cite{wang2025smamba}; DLinear~\cite{zeng2023dlinear} and TimeMixer~\cite{wang2024timemixer}; convolution-based ConvTimeNet~\cite{cheng2025convtimenet} and TimesNet~\cite{wu2023timesnet}; and Transformer-based PatchTST~\cite{nie2023patchtst}, iTransformer~\cite{liu2024itransformer}, and TimeXer~\cite{wang2024timexer}. These comparisons span statistical methods and diverse neural forecasting architectures. We additionally compare AIDE~\cite{jiang2025aide} and Optuna-TPE~\cite{akiba2019optuna,bergstra2011tpe}, both using a budget of 25 iterations. We report mean squared error (MSE) and mean absolute error (MAE); lower values are better. Main-table improvements use the best baseline for each dataset and metric. All component ablations use full NeoResearch as the reference across tasks.

\begin{figure}[t]
    \centering
        \includegraphics[width=\linewidth]{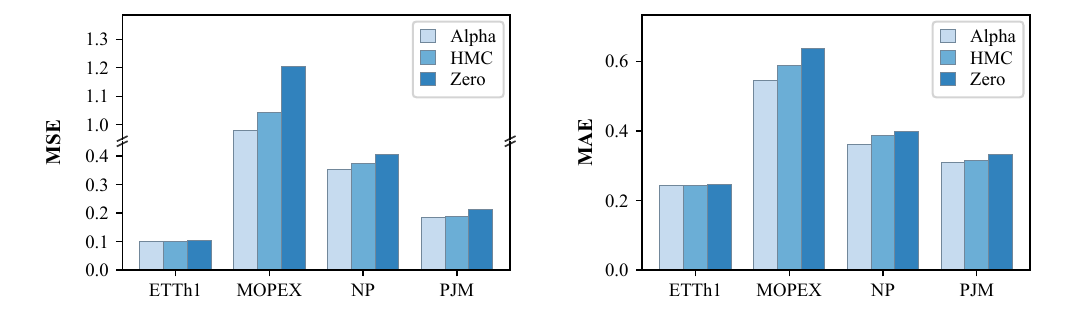}
            \caption{Research modes. Autonomous Alpha outperforms HMC and Zero on all four tasks.}
        \Description{Alpha has the lowest MSE and MAE on all four tasks, followed by HMC and then Zero.}
        \label{fig:research-variants}
\end{figure}

\subsubsection{Implementation Details}
We use DeepSeek-V4.1Flash as the default reasoning engine, with temperature 0.2, a maximum output length of 32k tokens per call, and up to three retries per request. Appendix~\ref{app:reasoning-backbones} compares alternative reasoning backbones. The default Alpha mode starts from a human-designed reference and proceeds autonomously for at most $k=25$ research iterations, stopping after seven consecutive rounds without improvement. Candidate forecasting models contain at most 10 million parameters; computation uses a single NVIDIA RTX 4090 GPU. Each candidate's recipe specifies its optimizer, loss, learning rate, regularization, batch size, seed, and training budget. Validation MSE determines checkpoint selection, early stopping, and final candidate selection. Test observations are excluded from these decisions and validation folds. Each target uses memory from cold-start research on the other nine datasets, as detailed in Appendix~\ref{app:memory-evaluation}; target evaluations do not form a transfer sequence. Section~\ref{sec:sequential-research-accumulation} evaluates sequential accumulation separately.

\subsection{Main Results}
\label{sec:main-results}
As shown in Table~\ref{tab:main-results}, we evaluate the held-out performance of NeoResearch's selected models to assess autonomous model discovery across domains. NeoResearch achieves the best results in most comparisons against representative statistical and neural forecasting methods. The strongest baseline varies across tasks, indicating that no single evaluated architecture is uniformly preferable. This variation supports adapting model design to the target data, as proposed in the outer discovery loop, where domain evidence informs choices of preprocessing, temporal modeling, and variable interaction. The comparisons with AIDE and Optuna-TPE provide a complementary assessment of the complete discovery procedure. NeoResearch consistently returns more accurate forecasters than these automated alternatives, supporting the workflow in Section~\ref{sec:research-cycle}, which combines literature- and experience-guided architecture exploration with iterative training optimization through experimental feedback. The discovered models perform numerical prediction independently of the LLM. Although candidates are selected using validation MSE, their advantage also extends to test MAE. Established baselines remain close on some tasks, and Windy presents a clear exception to the overall ranking: NeoResearch achieves the lowest MAE, while DLinear and DeepAR has lower MSE.

\subsection{Ablation Study}
\label{sec:ablation-studies}

\subsubsection{Dual-Loop Ablation}
\label{sec:evidence-analysis}
To assess the dual-loop workflow, we compare full NeoResearch with the two variants in Table~\ref{tab:evidence-ablation}. The w/o Outer loop variant fixes the design of the first new model generated under Alpha from the human-designed reference and retains training optimization; w/o Inner loop retains architecture discovery but evaluates each candidate once under a fixed training configuration, without iterative tuning. Full NeoResearch yields lower errors throughout, supporting the coordination of architecture discovery and training optimization. Inner-loop feedback informs whether to continue tuning or revise the architectural hypothesis. The ranking of the reduced workflows varies across datasets: w/o Outer loop generally performs worse, but ETTh1 favors it over w/o Inner loop.

\subsubsection{Hierarchical Memory Ablation}
\label{sec:dual-loop-analysis}
We use the ablations in Table~\ref{tab:memory-ablation} to test whether historical experience provides guidance beyond current experimental feedback. The three variants block historical architecture-level experience, historical optimization-level experience, or both levels of experience, while retaining both research loops and feedback from the current experiment. All restrictions worsen forecasting, supporting the reuse of architectural outcomes and tuning experience for hypothesis selection and training decisions. On PJM, however, removing architecture-level memory alone produces a higher MSE than removing both levels.

\begin{table}[t]
    \centering
    \caption{Knowledge extraction ablation. NeoResearch leads on all four tasks. Best results are bold; second-best results are underlined.}
    \label{tab:knowledge-ablation}
    \setlength{\tabcolsep}{1.5pt}
    \renewcommand{\arraystretch}{0.85}
    \resizebox{\linewidth}{!}{%
    \begin{tabular}{@{}lrrrrrrrr@{}}
    \toprule
    Method & \multicolumn{2}{c}{ETTh1} & \multicolumn{2}{c}{MOPEX} & \multicolumn{2}{c}{NP} & \multicolumn{2}{c}{PJM} \\
    \cmidrule(lr){2-3} \cmidrule(lr){4-5} \cmidrule(lr){6-7} \cmidrule(lr){8-9}
     & MSE & MAE & MSE & MAE & MSE & MAE & MSE & MAE \\
    \midrule
    w/o Extraction & 0.1043 & 0.2479 & 1.0378 & 0.6027 & \underline{0.3934} & \underline{0.3851} & \underline{0.1897} & \underline{0.3135} \\
    w/o Knowledge & \underline{0.1038} & \underline{0.2474} & \underline{1.0258} & \underline{0.5683} & 0.4212 & 0.4075 & 0.1969 & 0.3194 \\
    \midrule
    \textbf{NeoResearch} & \textbf{0.1004} & \textbf{0.2434} & \textbf{0.9812} & \textbf{0.5443} & \textbf{0.3546} & \textbf{0.3618} & \textbf{0.1849} & \textbf{0.3093} \\
    \bottomrule
    \end{tabular}%
    }
\end{table}

\begin{figure*}[t]
    \centering
    \includegraphics[width=\textwidth]{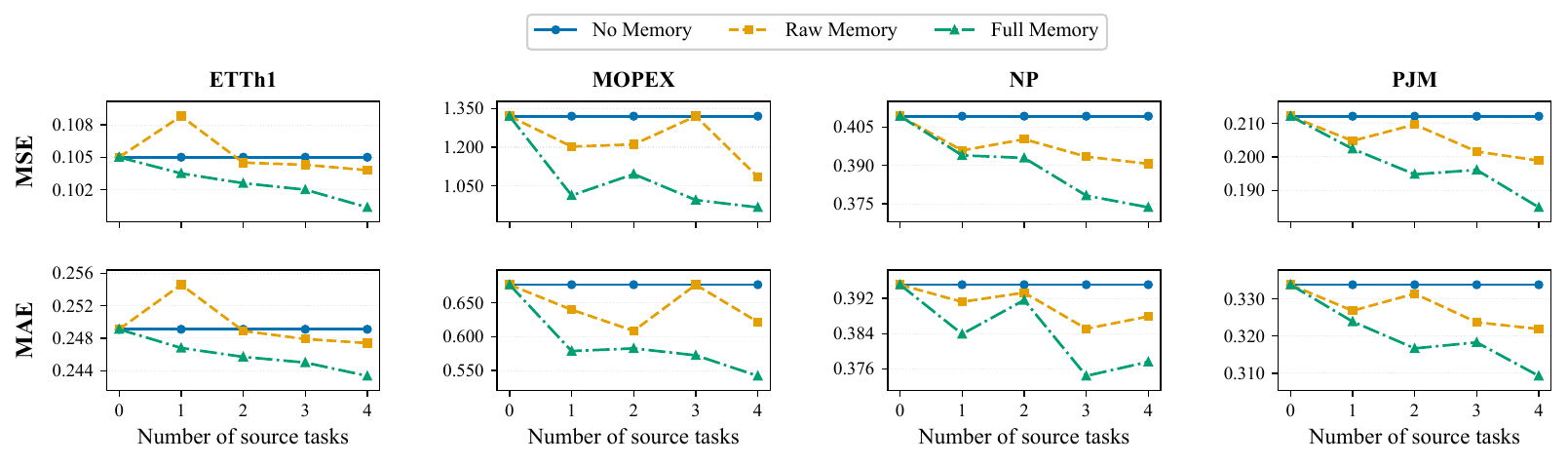}
        \caption{Target-task performance with accumulated source-task memory. Columns show target tasks; the horizontal axis counts source tasks added in order: GEF-L, Windy, BE, and FR. Zero denotes no source-task memory.}
    \Description{Four columns show ETTh1, MOPEX, NP, and PJM. The top row reports MSE and the bottom row reports MAE, comparing No Memory, Raw Memory, and Full Memory. Iteration zero denotes no source-task memory; iterations one through four cumulatively add GEF-L, Windy, BE, and FR. These checkpoints count source datasets, not research rounds within a task.}
    \label{fig:sequential-research-accumulation}

\end{figure*}

\subsubsection{Knowledge Extraction Ablation}
\label{sec:knowledge-ablation}
The comparison in Table~\ref{tab:knowledge-ablation} examines whether structured extraction makes literature more useful for architecture discovery. The w/o Extraction variant uses original text from newly retrieved papers and directly indexes the original sources for retained literature knowledge. The w/o Knowledge variant disables both sources while preserving experience memory and model assets. Full NeoResearch performs best throughout, supporting the extraction of mechanisms, evidence, and applicability conditions before literature is used to formulate architectural hypotheses. On MOPEX, however, direct source-text access performs worse than excluding literature.

\subsubsection{Cross-Task Reuse Ablation}
\label{sec:cross-task-memory}
The resource ablations in Figure~\ref{fig:cross-dataset-evolution} test whether reusing knowledge, experience, and model designs from prior tasks improves target-task discovery. Restricting any resource worsens forecasting across the evaluated tasks, even though new literature retrieval and within-task experience remain available. These results support retaining prior findings to inform architectural hypotheses and adapt candidate designs, as specified in Sections~\ref{sec:architecture-discovery} and~\ref{sec:research-memory}. Reused designs are trained with newly initialized weights and validated on the target task. The relative contribution of each resource varies, however. MOPEX is most sensitive to removing cross-task experience, whereas the most damaging restriction elsewhere depends on the task and error criterion.

\begin{figure}[t]
    \centering
        \includegraphics[width=\linewidth]{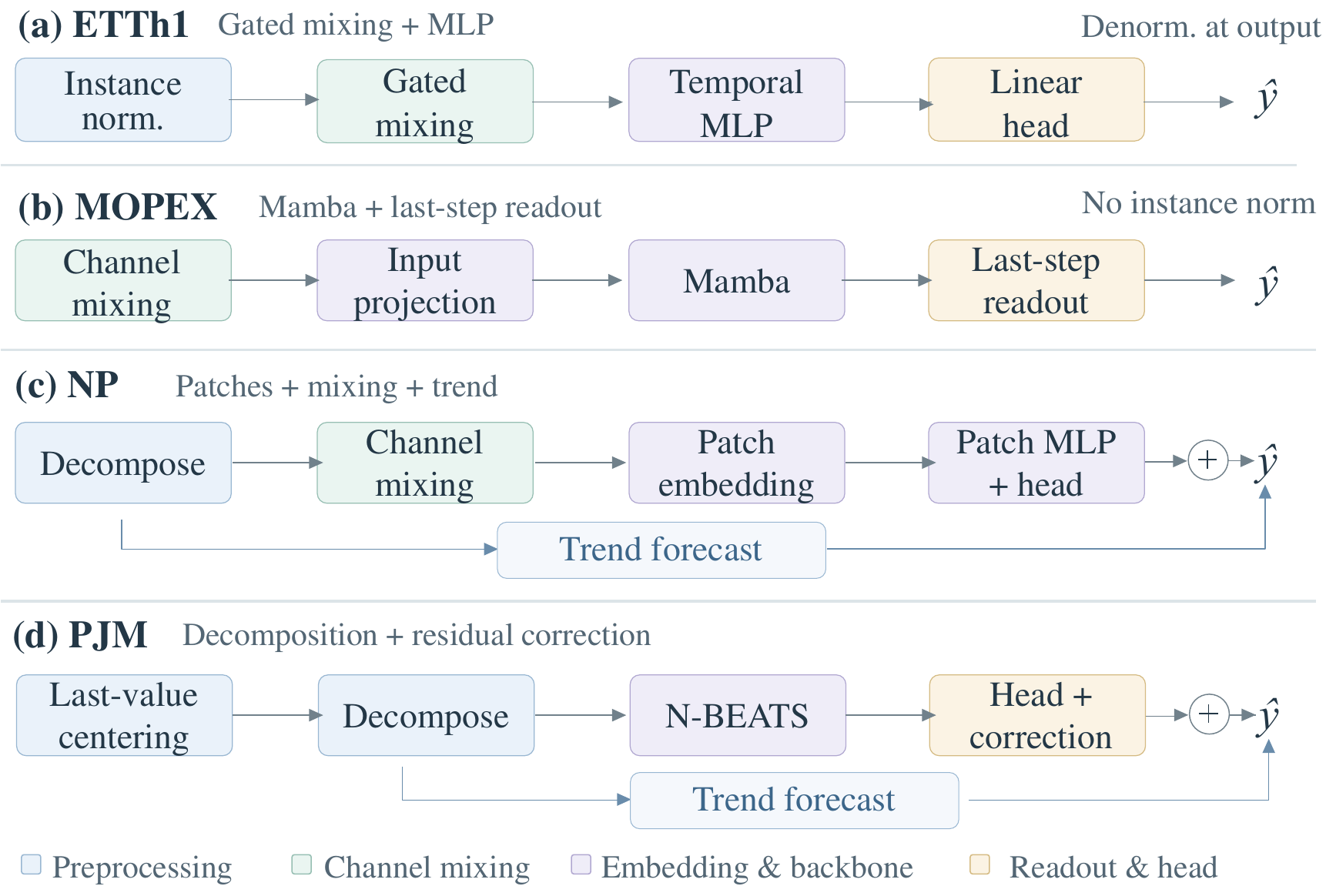}
            \caption{Discovered forecasting models vary in preprocessing, variable interaction, temporal modeling, and readout.}
        \Description{ETTh1 combines instance normalization, gated mixing, and a temporal MLP. MOPEX uses channel mixing, Mamba, and last-step readout without instance normalization. NP and PJM add trend forecasts to patch-based and N-BEATS prediction paths, respectively.}
        \label{fig:artifacts-analysis}
\end{figure}

\subsection{Exploration Study}
\label{sec:exploration-analysis}

\subsubsection{Model Design Analysis}
The models in Figure~\ref{fig:artifacts-analysis} vary in preprocessing, variable interaction, temporal modeling, and readout. Some combine learned prediction paths with explicit trend forecasts; others use a single temporal representation. ETTh1 combines gated mixing with a temporal MLP, while MOPEX couples channel mixing with Mamba and last-step readout. These task-specific designs reflect outer-loop exploration: evidence motivates hypotheses, while validation errors and convergence diagnostics guide architecture revision. MOPEX omits normalization despite observed distribution shift. As Section~\ref{sec:research-trajectory} shows, experiments overturned normalization's expected benefit, prompting revision of the preprocessing hypothesis and model recipe.

\begin{figure}[t]
    \centering
        \includegraphics[width=\linewidth]{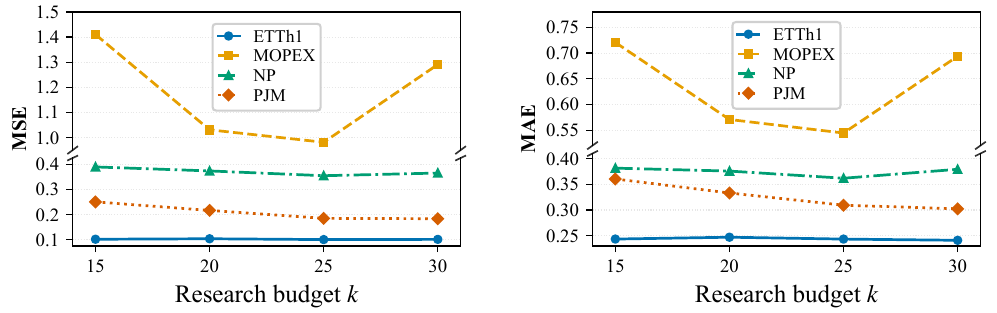}
            \caption{Research-budget sensitivity. PJM benefits most consistently; other tasks show non-monotonic gains.}
        \Description{PJM improves with budget, ETTh1 varies narrowly, and NP and MOPEX reach their lowest errors at 25 before worsening at 30.}
        \label{fig:research-budget}
    \medskip
    \centering
    \includegraphics[width=\linewidth]{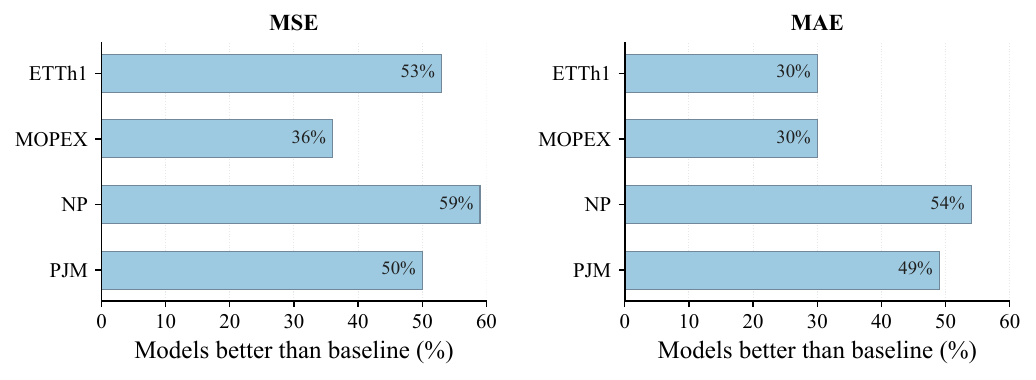}
        \caption{Quality of discovered candidates. NP has the largest share outperforming the baseline under each metric.}
    \Description{MSE yields are 53, 36, 59, and 50 percent for ETTh1, MOPEX, NP, and PJM; MAE yields are 30, 30, 54, and 49 percent.}
    \label{fig:model-factory-scaling}
\end{figure}

\begin{figure*}[t]
    \centering
    \includegraphics[width=0.96\textwidth]{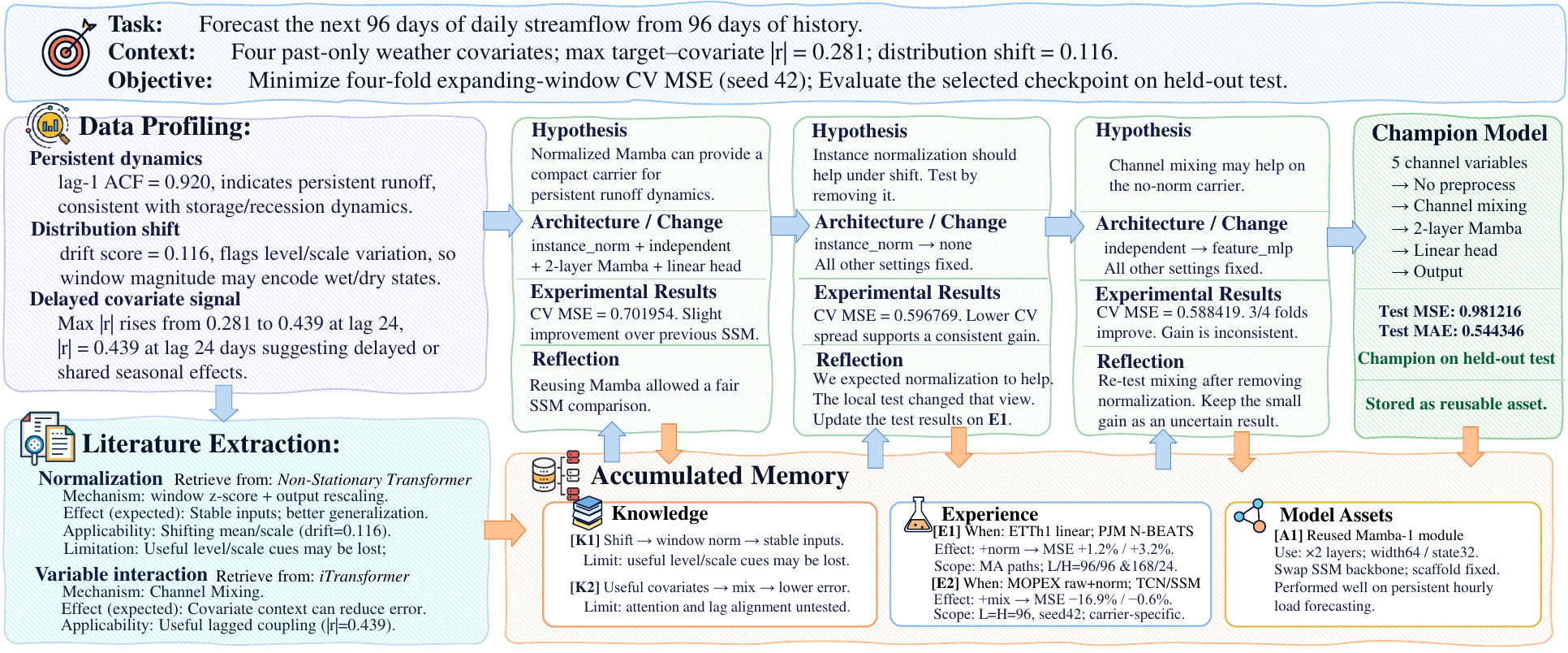}
        \caption{MOPEX research trajectory. Literature mechanisms and accumulated experience motivate architecture hypotheses; target-task experiments revise their applicability and produce a reusable forecasting model.}
    \label{fig:research-trajectory}
    \Description{Data profiling, literature knowledge, prior experience, and a reusable Mamba asset guide successive hypotheses. Experiments overturn the expected normalization benefit, retain the smaller mixing gain as uncertain, and save the champion recipe as a reusable asset.}
\end{figure*}

\subsubsection{Candidate Quality Analysis}
As shown in Figure~\ref{fig:model-factory-scaling}, we examine candidate pools beyond the selected models to assess whether discovery produces competitive alternatives. More than 100 candidates are evaluated per task, and every pool contains models outperforming the reference. However, many proposals do not improve on it, and candidate yield varies by task and metric. These outcomes support validation-based selection in the dual-loop workflow: architectural hypotheses must be implemented, trained, and evaluated before their value can be established. MOPEX illustrates why final-model quality and candidate yield should be distinguished, combining a large final-model gain with a relatively small share of competitive candidates.

\subsubsection{Research Mode Analysis}
Figure~\ref{fig:research-variants} compares modes with different initialization and continued human involvement. Alpha leads throughout, followed by Human-Machine Cooperation (HMC) and Zero. Zero starts without a reference or human interaction, whereas Alpha proceeds autonomously from a human-designed reference; both stop after seven consecutive rounds without improvement. The ranking favors reference-assisted autonomous discovery, in which experimental feedback guides subsequent architecture hypotheses. HMC shares Alpha's reference but requires expert guidance at this plateau before continuing within the remaining budget. It does not surpass Alpha, although this comparison concerns complete research modes: HMC changes both expert interaction and the continuation rule.

\subsubsection{Research Budget Analysis}
\label{sec:budget-sensitivity}
To examine whether longer investigations improve forecasting, we compare maximum budgets of $k\in\{15,20,25,30\}$ in Figure~\ref{fig:research-budget}. Additional iterations do not consistently reduce test error: some tasks improve before worsening, while others change little. Architecture exploration and training optimization share the research budget, so further rounds permit more hypotheses and training adjustments without guaranteeing a better returned model. Each point represents a separate investigation under its own maximum budget, rather than a checkpoint of one trajectory. Final selection uses validation MSE, and the plotted test errors assess each investigation's outcome. PJM shows the clearest sustained benefit from increasing the budget.

\subsubsection{Memory Structure and Accumulation Analysis}
\label{sec:sequential-research-accumulation}
Figure~\ref{fig:sequential-research-accumulation} compares No Memory, Raw Memory, and Full Memory to assess how structured memory supports discovery as source tasks accumulate. All variants share source histories and fixed target budgets. Raw Memory retains research resources without formalization, while Full Memory organizes them with supporting evidence and applicability conditions. Full Memory outperforms both alternatives at every nonzero checkpoint and generally improves as sources accumulate. This supports consolidating research outcomes into conditional lessons that guide architecture discovery and training optimization. On MOPEX, however, the second source temporarily worsens Full Memory's performance before later additions improve it. This local reversal reinforces the need to reassess transferred findings against target-task evidence.

\subsection{Case Study}
\label{sec:research-trajectory}
Figure~\ref{fig:research-trajectory} traces streamflow discovery using literature, experience, and experimental feedback. Persistent runoff and delayed covariate signals motivate Mamba reuse and variable-interaction tests. Preprocessing revision yields the largest validation gain; backbone replacement and channel mixing yield smaller gains. Removing normalization unexpectedly improves performance across all four expanding-window validation folds under distribution shift. Mixing improves mean validation performance, but inconsistent fold-level gains leave its benefit uncertain. Memory retains the selected recipe and these findings with task-specific applicability conditions. Held-out testing follows selection by validation MSE.
\label{page:experiments-end}
\par

\section{Discussion}
\label{sec:discussion}

NeoResearch uses literature to formulate testable hypotheses rather than design prescriptions. As illustrated by the MOPEX case study, experimental evidence can challenge literature-informed assumptions and refine their applicability.  Meanwhile, the memory analyses demonstrate that structured, condition-aware memory supports cumulative research across stages and tasks, although more experience does not always guarantee better outcomes. These findings favor hypothesis revision and selective experience reuse over larger knowledge repositories. Discovery is constrained by architecture space and research budget, with its strongest results benefiting from human-designed reference models. Future work will explore fully cold-start discovery, robustness to conflicting scientific evidence, and more reliable, efficient cross-domain experience transfer.

\section{Conclusion}
\label{sec:conclusion}

We presented NeoResearch, a literature- and experience-guided framework for discovering lightweight, domain-specific time series forecasting models. Its dual-loop workflow coordinates architecture exploration and training optimization, while autonomous literature mining and structured research memory support knowledge-informed decisions and experience reuse across research stages and tasks. Experiments on ten datasets show competitive performance and benefits from knowledge and experience. These results support cumulative model discovery: knowledge motivates hypotheses, experiments assess applicability, and lessons guide subsequent investigations. Future work will improve exploration efficiency and reliability of cross-domain experience transfer.

\bibliographystyle{ACM-Reference-Format-no-links}
\bibliography{main}

@article{lago2021epf,
  author  = {Jesus Lago and Grzegorz Marcjasz and Bart De Schutter and Rafa{\l} Weron},
  title   = {Forecasting day-ahead electricity prices: A review of state-of-the-art algorithms, best practices and an open-access benchmark},
  journal = {Applied Energy},
  volume  = {293},
  pages   = {116983},
  year    = {2021},
  doi     = {10.1016/j.apenergy.2021.116983}
}

@inproceedings{zhou2021informer,
  author    = {Haoyi Zhou and Shanghang Zhang and Jieqi Peng and Shuai Zhang and Jianxin Li and Hui Xiong and Wancai Zhang},
  title     = {{Informer}: Beyond Efficient Transformer for Long Sequence Time-Series Forecasting},
  booktitle = {Proceedings of the AAAI Conference on Artificial Intelligence},
  volume    = {35},
  number    = {12},
  pages     = {11106--11115},
  year      = {2021},
  doi       = {10.1609/aaai.v35i12.17325}
}

@article{hong2016gefcom2014,
  author  = {Tao Hong and Pierre Pinson and Shu Fan and Hamidreza Zareipour and Alberto Troccoli and Rob J. Hyndman},
  title   = {Probabilistic energy forecasting: {Global Energy Forecasting Competition 2014} and beyond},
  journal = {International Journal of Forecasting},
  volume  = {32},
  number  = {3},
  pages   = {896--913},
  year    = {2016},
  doi     = {10.1016/j.ijforecast.2016.02.001}
}

@incollection{schaake2006mopex,
  author    = {John Schaake and Shuzheng Cong and Qingyun Duan},
  title     = {The {US MOPEX} Data Set},
  booktitle = {Large Sample Basin Experiments for Hydrological Model Parameterization: Results of the Model Parameter Experiment--MOPEX},
  series    = {IAHS Publication},
  volume    = {307},
  pages     = {9--28},
  publisher = {International Association of Hydrological Sciences},
  year      = {2006},
  url       = {https://iahs.info/uploads/dms/13600.04-9-28-SCHAAKE.pdf}
}

@inproceedings{zeng2023dlinear,
  author    = {Ailing Zeng and Muxi Chen and Lei Zhang and Qiang Xu},
  title     = {Are Transformers Effective for Time Series Forecasting?},
  booktitle = {Proceedings of the AAAI Conference on Artificial Intelligence},
  volume    = {37},
  number    = {9},
  pages     = {11121--11128},
  year      = {2023},
  doi       = {10.1609/aaai.v37i9.26317},
  url       = {https://ojs.aaai.org/index.php/AAAI/article/view/26317}
}

@inproceedings{nie2023patchtst,
  author    = {Yuqi Nie and Nam H. Nguyen and Phanwadee Sinthong and Jayant Kalagnanam},
  title     = {A Time Series is Worth 64 Words: Long-term Forecasting with Transformers},
  booktitle = {The Eleventh International Conference on Learning Representations},
  year      = {2023},
  url       = {https://openreview.net/forum?id=Jbdc0vTOcol}
}

@inproceedings{liu2024itransformer,
  author    = {Yong Liu and Tengge Hu and Haoran Zhang and Haixu Wu and
               Shiyu Wang and Lintao Ma and Mingsheng Long},
  title     = {{iTransformer}: Inverted Transformers Are Effective for Time Series Forecasting},
  booktitle = {The Twelfth International Conference on Learning Representations},
  year      = {2024},
  url       = {https://openreview.net/forum?id=JePfAI8fah}
}

@inproceedings{jin2024timellm,
  author    = {Ming Jin and Shiyu Wang and Lintao Ma and Zhixuan Chu and
               James Y. Zhang and Xiaoming Shi and Pin-Yu Chen and
               Yuxuan Liang and Yuan-Fang Li and Shirui Pan and Qingsong Wen},
  title     = {{Time-LLM}: Time Series Forecasting by Reprogramming Large Language Models},
  booktitle = {The Twelfth International Conference on Learning Representations},
  year      = {2024},
  url       = {https://openreview.net/forum?id=Unb5CVPtae}
}

@misc{lu2024aiscientist,
  author        = {Chris Lu and Cong Lu and Robert Tjarko Lange and
                   Jakob Foerster and Jeff Clune and David Ha},
  title         = {The {AI} Scientist: Towards Fully Automated Open-Ended Scientific Discovery},
  year          = {2024},
  eprint        = {2408.06292},
  archivePrefix = {arXiv},
  primaryClass  = {cs.AI},
  url           = {https://arxiv.org/abs/2408.06292v3}
}

@article{ye2026tsreasoner,
  author  = {Wen Ye and Wei Yang and Defu Cao and Yizhou Zhang and
             Lumingyuan Tang and Jie Cai and Yan Liu},
  title   = {{TS-Reasoner}: Domain-Oriented Time Series Inference Agents for Reasoning and Automated Analysis},
  journal = {Transactions on Machine Learning Research},
  year    = {2026},
  url     = {https://openreview.net/forum?id=yhy7Vigjcf}
}

@article{wang2024autots,
  author  = {Chunnan Wang and Xingyu Chen and Chengyue Wu and Hongzhi Wang},
  title   = {Automatic Time Series Forecasting Model Design Based on Pruning},
  journal = {Applied Soft Computing},
  volume  = {162},
  pages   = {111804},
  year    = {2024},
  doi     = {10.1016/j.asoc.2024.111804},
  url     = {https://www.sciencedirect.com/science/article/abs/pii/S1568494624005787}
}

@inproceedings{zhang2023autostl,
  author    = {Zijian Zhang and Xiangyu Zhao and Hao Miao and Chunxu Zhang and
               Hongwei Zhao and Junbo Zhang},
  title     = {{AutoSTL}: Automated Spatio-Temporal Multi-Task Learning},
  booktitle = {Proceedings of the AAAI Conference on Artificial Intelligence},
  volume    = {37},
  pages     = {4902--4910},
  publisher = {AAAI Press},
  year      = {2023},
  doi       = {10.1609/aaai.v37i4.25616},
  url       = {https://ojs.aaai.org/index.php/AAAI/article/view/25616}
}

@inproceedings{wang2024scimon,
  author    = {Qingyun Wang and Doug Downey and Heng Ji and Tom Hope},
  title     = {{SciMON}: Scientific Inspiration Machines Optimized for Novelty},
  booktitle = {Proceedings of the 62nd Annual Meeting of the Association for Computational Linguistics (Volume 1: Long Papers)},
  pages     = {279--299},
  address   = {Bangkok, Thailand},
  publisher = {Association for Computational Linguistics},
  year      = {2024},
  doi       = {10.18653/v1/2024.acl-long.18},
  url       = {https://aclanthology.org/2024.acl-long.18/}
}

@inproceedings{baek2025researchagent,
  author    = {Jinheon Baek and Sujay Kumar Jauhar and Silviu Cucerzan and Sung Ju Hwang},
  title     = {{ResearchAgent}: Iterative Research Idea Generation over Scientific Literature with Large Language Models},
  booktitle = {Proceedings of the 2025 Conference of the Nations of the Americas Chapter of the Association for Computational Linguistics: Human Language Technologies (Volume 1: Long Papers)},
  pages     = {6709--6738},
  address   = {Albuquerque, New Mexico},
  publisher = {Association for Computational Linguistics},
  year      = {2025},
  doi       = {10.18653/v1/2025.naacl-long.342},
  url       = {https://aclanthology.org/2025.naacl-long.342/}
}

@inproceedings{gruver2023llmtime,
  author    = {Nate Gruver and Marc Finzi and Shikai Qiu and Andrew Gordon Wilson},
  title     = {Large Language Models Are Zero-Shot Time Series Forecasters},
  booktitle = {Advances in Neural Information Processing Systems},
  volume    = {36},
  year      = {2023},
  url       = {https://arxiv.org/abs/2310.07820}
}

@inproceedings{zhou2026timer1,
  author    = {Yitong Zhou and Yucong Luo and Mingyue Cheng and Qi Liu and
               Jiahao Wang and Daoyu Wang and Enhong Chen},
  title     = {Time Series Forecasting via Reasoning: A Slow-Thinking Approach
               with Reinforcement Fine-Tuned {LLMs}},
  booktitle = {Proceedings of the 35th ACM International Conference on
               Information and Knowledge Management},
  address   = {Rome, Italy},
  publisher = {Association for Computing Machinery},
  year      = {2026},
  doi       = {10.1145/3799682.3840737},
  url       = {https://arxiv.org/html/2506.10630v4}
}

@misc{cao2025conversational,
  author        = {Defu Cao and Michael Gee and Jinbo Liu and Hengxuan Wang and
                   Wei Yang and Rui Wang and Yan Liu},
  title         = {Conversational Time Series Foundation Models: Towards Explainable and Effective Forecasting},
  year          = {2025},
  eprint        = {2512.16022},
  archivePrefix = {arXiv},
  primaryClass  = {cs.AI},
  url           = {https://arxiv.org/abs/2512.16022v1}
}

@inproceedings{tao2026memcast,
  author    = {Xiaoyu Tao and Mingyue Cheng and Ze Guo and Shuo Yu and Yaguo Liu and Qi Liu and Shijin Wang},
  title     = {{MemCast}: Memory-Driven Time Series Forecasting with Experience-Conditioned Reasoning},
  booktitle = {Proceedings of the 43rd International Conference on Machine Learning},
  series    = {Proceedings of Machine Learning Research},
  volume    = {306},
  pages     = {119553--119569},
  publisher = {PMLR},
  year      = {2026},
  url       = {https://proceedings.mlr.press/v306/tao26a.html}
}

@inproceedings{yang2025nader,
  author    = {Zekang Yang and Wang Zeng and Sheng Jin and Chen Qian and Ping Luo and Wentao Liu},
  title     = {{NADER}: Neural Architecture Design via Multi-Agent Collaboration},
  booktitle = {Proceedings of the IEEE/CVF Conference on Computer Vision and Pattern Recognition (CVPR)},
  pages     = {4452--4461},
  month     = jun,
  year      = {2025},
  url       = {https://openaccess.thecvf.com/content/CVPR2025/html/Yang_NADER_Neural_Architecture_Design_via_Multi-Agent_Collaboration_CVPR_2025_paper.html}
}

@inproceedings{tang2025airesearcher,
  author    = {Tang, Jiabin and Xia, Lianghao and Li, Zhonghang and Huang, Chao},
  title     = {{AI-Researcher}: Autonomous Scientific Innovation},
  booktitle = {Advances in Neural Information Processing Systems},
  volume    = {38},
  year      = {2025},
  doi       = {10.52202/085713-0320},
  url       = {https://doi.org/10.52202/085713-0320}
}

@inproceedings{jansen2025codescientist,
  title     = {{CodeScientist}: End-to-End Semi-Automated Scientific Discovery with Code-based Experimentation},
  author    = {Jansen, Peter and Tafjord, Oyvind and Radensky, Marissa and Siangliulue, Pao and Hope, Tom and Dalvi Mishra, Bhavana and Majumder, Bodhisattwa Prasad and Weld, Daniel S. and Clark, Peter},
  booktitle = {Findings of the Association for Computational Linguistics: ACL 2025},
  month     = jul,
  year      = {2025},
  address   = {Vienna, Austria},
  publisher = {Association for Computational Linguistics},
  url       = {https://doi.org/10.18653/v1/2025.findings-acl.692},
  doi       = {10.18653/v1/2025.findings-acl.692},
  pages     = {13370--13467},
  isbn      = {979-8-89176-256-5}
}

@misc{zhao2025timeseriesscientist,
  title         = {{TimeSeriesScientist}: A General-Purpose {AI} Agent for Time Series Analysis},
  author        = {Zhao, Haokun and Zhang, Xiang and Wei, Jiaqi and Xu, Yiwei and He, Yuting and Sun, Siqi and You, Chenyu},
  year          = {2025},
  eprint        = {2510.01538v2},
  archivePrefix = {arXiv},
  primaryClass  = {cs.LG},
  url           = {https://arxiv.org/abs/2510.01538v2}
}

@inproceedings{trirat2025monaq,
  title     = {{MONAQ}: Multi-Objective Neural Architecture Querying for Time-Series Analysis on Resource-Constrained Devices},
  author    = {Trirat, Patara and Lee, Jae-Gil},
  booktitle = {Findings of the Association for Computational Linguistics: EMNLP 2025},
  address   = {Suzhou, China},
  month     = nov,
  year      = {2025},
  publisher = {Association for Computational Linguistics},
  pages     = {16922--16950},
  doi       = {10.18653/v1/2025.findings-emnlp.918},
  url       = {https://aclanthology.org/2025.findings-emnlp.918/}
}

@misc{shen2026metacaster,
  title         = {{MetaCaster}: Meta-Harness-Optimized Agent for End-to-End Few-Shot Learning of Lightweight Time Series Forecasters},
  author        = {Shen, ChengAo and Yu, Wenchao and Wu, Fangyu and Song, Dongjin and Tong, Hanghang and Luo, Dongsheng and Cheng, Wei and Chen, Haifeng and Ni, Jingchao},
  year          = {2026},
  eprint        = {2608.23473v2},
  archivePrefix = {arXiv},
  primaryClass  = {cs.LG},
  url           = {https://arxiv.org/abs/2608.23473v2}
}

@misc{li2025rdagentquant,
  title         = {{R\&D-Agent-Quant}: A Multi-Agent Framework for Data-Centric Factors and Model Joint Optimization},
  author        = {Li, Yuante and Yang, Xu and Yang, Xiao and Xu, Minrui and Wang, Xisen and Liu, Weiqing and Bian, Jiang},
  year          = {2025},
  eprint        = {2505.15155v2},
  archivePrefix = {arXiv},
  primaryClass  = {q-fin.CP},
  url           = {https://arxiv.org/abs/2505.15155v2}
}

@misc{lyu2026evoscientist,
  title         = {{EvoScientist}: Towards Multi-Agent Evolving {AI} Scientists for End-to-End Scientific Discovery},
  author        = {Lyu, Yougang and Zhang, Xi and Yi, Xinhao and Zhao, Yuyue and Guo, Shuyu and Hu, Wenxiang and Piotrowski, Jan and Kaliski, Jakub and Urbani, Jacopo and Meng, Zaiqiao and Zhou, Lun and Yan, Xiaohui},
  year          = {2026},
  eprint        = {2603.08127v1},
  archivePrefix = {arXiv},
  primaryClass  = {cs.CL},
  url           = {https://arxiv.org/abs/2603.08127v1}
}

@misc{yang2025rdagent,
  title         = {{R\&D-Agent}: An {LLM}-Agent Framework Towards Autonomous Data Science},
  author        = {Yang, Xu and Yang, Xiao and Fang, Shikai and Zhang, Yifei and Wang, Jian and Xian, Bowen and Li, Qizheng and Li, Jingyuan and Xu, Minrui and Li, Yuante and Pan, Haoran and Zhang, Yuge and Liu, Weiqing and Shen, Yelong and Chen, Weizhu and Bian, Jiang},
  year          = {2025},
  eprint        = {2505.14738v2},
  archiveprefix = {arXiv},
  primaryclass  = {cs.AI},
  url           = {https://arxiv.org/abs/2505.14738v2}
}

@book{box2015arima,
  author    = {George E. P. Box and Gwilym M. Jenkins and Gregory C. Reinsel and Greta M. Ljung},
  title     = {Time Series Analysis: Forecasting and Control},
  edition   = {5},
  publisher = {Wiley},
  year      = {2015},
  isbn      = {9781118675021}
}

@article{taylor2018prophet,
  author  = {Sean J. Taylor and Benjamin Letham},
  title   = {Forecasting at Scale},
  journal = {The American Statistician},
  volume  = {72},
  number  = {1},
  pages   = {37--45},
  year    = {2018},
  doi     = {10.1080/00031305.2017.1380080}
}

@article{elman1990rnn,
  author  = {Jeffrey L. Elman},
  title   = {Finding Structure in Time},
  journal = {Cognitive Science},
  volume  = {14},
  number  = {2},
  pages   = {179--211},
  year    = {1990},
  doi     = {10.1207/s15516709cog1402_1}
}

@article{hochreiter1997lstm,
  author  = {Sepp Hochreiter and J{\"u}rgen Schmidhuber},
  title   = {Long Short-Term Memory},
  journal = {Neural Computation},
  volume  = {9},
  number  = {8},
  pages   = {1735--1780},
  year    = {1997},
  doi     = {10.1162/neco.1997.9.8.1735}
}

@inproceedings{wu2020mtgnn,
  author = {Wu, Zonghan and Pan, Shirui and Long, Guodong and Jiang, Jing and Chang, Xiaojun and Zhang, Chengqi},
  title = {Connecting the Dots: Multivariate Time Series Forecasting with Graph Neural Networks},
  booktitle = {Proceedings of the 26th ACM SIGKDD International Conference on Knowledge Discovery \& Data Mining},
  year = {2020},
  pages = {753--763},
  publisher = {Association for Computing Machinery},
  doi = {10.1145/3394486.3403118},
  url = {https://doi.org/10.1145/3394486.3403118}
}

@article{salinas2020deepar,
  author = {Salinas, David and Flunkert, Valentin and Gasthaus, Jan and Januschowski, Tim},
  title = {{DeepAR}: Probabilistic Forecasting with Autoregressive Recurrent Networks},
  journal = {International Journal of Forecasting},
  year = {2020},
  volume = {36},
  number = {3},
  pages = {1181--1191},
  doi = {10.1016/j.ijforecast.2019.07.001},
  url = {https://doi.org/10.1016/j.ijforecast.2019.07.001}
}

@article{scarselli2009gnn,
  author = {Scarselli, Franco and Gori, Marco and Tsoi, Ah Chung and Hagenbuchner, Markus and Monfardini, Gabriele},
  title = {The Graph Neural Network Model},
  journal = {IEEE Transactions on Neural Networks},
  year = {2009},
  volume = {20},
  number = {1},
  pages = {61--80},
  doi = {10.1109/TNN.2008.2005605},
  url = {https://doi.org/10.1109/TNN.2008.2005605}
}

@inproceedings{gu2024mamba,
  author    = {Albert Gu and Tri Dao},
  title     = {{Mamba}: Linear-Time Sequence Modeling with Selective State Spaces},
  booktitle = {First Conference on Language Modeling},
  year      = {2024},
  url       = {https://openreview.net/forum?id=tEYskw1VY2}
}

@article{wang2025smamba,
  author  = {Zi-Han Wang and Fan-Heng Kong and Shi Feng and Ming Wang and Xiao-Cui Yang and Han Zhao and Da-Ling Wang and Yi-Fei Zhang},
  title   = {Is {Mamba} Effective for Time Series Forecasting?},
  journal = {Neurocomputing},
  volume  = {619},
  pages   = {129178},
  year    = {2025},
  doi     = {10.1016/j.neucom.2024.129178},
  url     = {https://www.sciencedirect.com/science/article/pii/S0925231224019490}
}

@inproceedings{cheng2025convtimenet,
  author    = {Mingyue Cheng and Jiqian Yang and Tingyue Pan and Qi Liu and Zhi Li and Shijin Wang},
  title     = {{ConvTimeNet}: A Deep Hierarchical Fully Convolutional Model for Multivariate Time Series Analysis},
  booktitle = {Companion Proceedings of the ACM on Web Conference 2025},
  series    = {WWW '25},
  pages     = {171--180},
  publisher = {Association for Computing Machinery},
  year      = {2025},
  doi       = {10.1145/3701716.3715214},
  url       = {https://doi.org/10.1145/3701716.3715214}
}

@inproceedings{wang2024timemixer,
  author    = {Shiyu Wang and Haixu Wu and Xiaoming Shi and Tengge Hu and
               Huakun Luo and Lintao Ma and James Y. Zhang and Jun Zhou},
  title     = {{TimeMixer}: Decomposable Multiscale Mixing for Time Series Forecasting},
  booktitle = {The Twelfth International Conference on Learning Representations},
  year      = {2024},
  url       = {https://openreview.net/forum?id=7oLshfEIC2}
}

@inproceedings{wu2023timesnet,
  author    = {Haixu Wu and Tengge Hu and Yong Liu and Hang Zhou and
               Jianmin Wang and Mingsheng Long},
  title     = {{TimesNet}: Temporal {2D}-Variation Modeling for General Time Series Analysis},
  booktitle = {The Eleventh International Conference on Learning Representations},
  year      = {2023},
  url       = {https://openreview.net/forum?id=ju_Uqw384Oq}
}

@inproceedings{wang2024timexer,
  author    = {Yuxuan Wang and Haixu Wu and Jiaxiang Dong and Guo Qin and
               Haoran Zhang and Yong Liu and Yunzhong Qiu and
               Jianmin Wang and Mingsheng Long},
  title     = {{TimeXer}: Empowering Transformers for Time Series Forecasting with Exogenous Variables},
  booktitle = {Advances in Neural Information Processing Systems},
  volume    = {37},
  year      = {2024},
  doi       = {10.52202/079017-0015},
  url       = {https://proceedings.neurips.cc/paper_files/paper/2024/hash/0113ef4642264adc2e6924a3cbbdf532-Abstract-Conference.html}
}

@misc{jiang2025aide,
  author        = {Zhengyao Jiang and Dominik Schmidt and Dhruv Srikanth and
                   Dixing Xu and Ian Kaplan and Deniss Jacenko and Yuxiang Wu},
  title         = {{AIDE}: {AI}-Driven Exploration in the Space of Code},
  year          = {2025},
  eprint        = {2502.13138},
  archivePrefix = {arXiv},
  primaryClass  = {cs.AI}
}

@inproceedings{akiba2019optuna,
  author    = {Takuya Akiba and Shotaro Sano and Toshihiko Yanase and
               Takeru Ohta and Masanori Koyama},
  title     = {{Optuna}: A Next-generation Hyperparameter Optimization Framework},
  booktitle = {Proceedings of the 25th ACM SIGKDD International Conference on Knowledge Discovery \& Data Mining},
  pages     = {2623--2631},
  year      = {2019}
}

@inproceedings{bergstra2011tpe,
  author    = {James S. Bergstra and R{\'e}mi Bardenet and Yoshua Bengio and
               Bal{\'a}zs K{\'e}gl},
  title     = {Algorithms for Hyper-Parameter Optimization},
  booktitle = {Advances in Neural Information Processing Systems},
  volume    = {24},
  pages     = {2546--2554},
  year      = {2011}
}
\appendix

\section{Dual-Loop Research Workflow and Execution}
\label{app:research-workflow}

A task specifies its dataset, timestamps, target, historical covariates, input length, forecast horizon, validation protocol, and resource limits. Diagnostics characterize missing values, distributions, temporal dynamics, and variable relationships. Test observations are excluded from diagnostics and research decisions.

\subsection{Outer-Loop Architecture Discovery}
\label{app:outer-loop}
The outer loop develops architectural hypotheses from task diagnostics, scientific literature, and architecture-level experience. Literature records retain source passages, modeling mechanisms, empirical evidence, applicability conditions, and limitations. The researcher checks the assumptions underlying published findings against the target data and forecasting objective. Each hypothesis states a proposed mechanism, its expected effect, its applicability conditions, and observations that could challenge it.

At iteration $i$, the selected hypothesis becomes a candidate design $d_i=(p_i,a_i)$, specifying input configuration, preprocessing, architecture, and module composition. The inner loop trains and validates this design. Its selected model $\hat{f}_i$ and experiment history $E_i$ support hypothesis assessment and the choice between further optimization, architectural refinement, or an alternative hypothesis. Before interpreting a result mechanistically, reflection checks implementation correctness, evaluation consistency, training adequacy, and confounding changes. Controlled comparisons name a reference model and declared intervention, holding other architecture and training settings fixed. Unresolved findings may motivate additional diagnostics or literature retrieval; negative or conflicting evidence remains linked to its experimental conditions.

\subsection{Inner-Loop Training Optimization}
\label{app:inner-loop}
For a fixed design $d_i$, the inner loop performs compatibility and preliminary execution checks, then iteratively trains and validates candidate models. Attempt $j$ executes recipe $r_{i,j}$ from Equation~\ref{eq:model-recipe}: preprocessing and architecture remain unchanged, while training loss $o_{i,j}$ and configuration $c_{i,j}$ may vary. Validation metrics, convergence curves, error analysis, and optimization-level experience guide adjustments to learning rate, batch size, regularization, scheduling, and loss. Implementation errors are repaired while preserving the hypothesis and experimental conditions. Changes to preprocessing, the backbone, or module composition require an outer-loop revision; incompatible designs also return to that loop.

Each record $e_{i,j}$ retains the executed recipe, training conditions, measurements, diagnostics, and execution status. The history $E_i$ includes completed and failed attempts, while $\mathcal{F}_i$ includes only successfully trained and validated models. Equation~\ref{eq:inner-selection} selects $\hat{f}_i$ by validation MSE regardless of the training loss. Final selection likewise minimizes validation MSE over all eligible candidates in $\mathcal{F}$. The selected model and full history return to the outer loop for assessment. Convergence, validation gains, and the shared budget determine whether further optimization is justified. Failures and plateaus support diagnosis without directly refuting the architecture in the current setting.

\subsection{Research Modes and Stopping}
\label{app:research-configuration}
The default budget is 25 iterations shared by both loops. Zero starts without a preset reference or human interaction; Alpha proceeds autonomously from a human-designed reference. Both stop after seven consecutive rounds without improvement. HMC uses Alpha's reference and requires expert guidance at that plateau before continuing within the remaining budget. A task-specified metric target, planner decision, or three consecutive duplicate proposals can also stop research. Training-level stopping applies to an individual attempt and is distinct from research-level stopping. Design revisions and resumed sessions retain consumed iterations, time, and resources across both loops.

\subsection{Recipes and Execution State}
\label{app:research-harness}
Table~\ref{tab:model-recipe-specification} summarizes the recipe components. A TypeScript coordinator invokes a Python backend for compatibility checks, preliminary execution, full training and validation, and held-out evaluation. Checks cover representations, outputs, branch composition, and the declared intervention. Only completed full evaluations support performance comparisons. Implementation repairs preserve the hypothesis and experimental conditions; incompatible designs return to the outer loop.

\begin{table}[t]
    \centering
    \caption{Model-recipe specification. The task fixes the forecasting interface and resource limits. The outer loop chooses $p_i$ and $a_i$; the inner loop varies $o_{i,j}$ and $c_{i,j}$ for that design.}
    \label{tab:model-recipe-specification}
    \begin{tabular}{@{}p{0.22\linewidth}p{0.73\linewidth}@{}}
        \toprule
        Field & Supported choices or constraints \\
        \midrule
        Input and preprocessing & Task-specified history, horizon, and observed-past covariates; normalization, decomposition, seasonal differencing, missingness masks, sample reweighting, last-value centering, or none. \\
        Embedding & Single/multiscale patches, inverted, frequency, calendar, or none. \\
        Temporal backbone & Linear, MLP, Transformer, recurrent, temporal convolution, DLinear, N-BEATS, mixer, gated depthwise convolution, TimesNet-lite, diagonal SSM, Mamba, or TimesBlock encoder. \\
        Variable interaction & Temporal, variate attention, temporal-then-feature, feature-then-temporal, graph, gated hybrid, or feature MLP. \\
        Layout and output & Channel-latent, sequence-latent, or channel-sequence layout; none/mean/last/flatten readout; compatible identity, linear, quantile, Gaussian, or residual-correction head for prediction. \\
        Composition & Single branch, gated parallel, or trend/residual branches; no nested composition. \\
        Training configuration & Loss, optimizer, learning rate, regularization, scheduling, batch size, seed, epoch/time limits, and early stopping, subject to task-level resource limits during model training. \\
        \bottomrule
    \end{tabular}
\end{table}

Persistent state $S$ links hypotheses, recipes, runs, artifacts, assessments, and the best model so far. Records preserve data and training settings, validation metrics, convergence curves, resource use, and execution status. Decision logs connect each intervention to its rationale and subsequent outcome, allowing unsuccessful attempts and unresolved questions to inform later choices. Curves retain phase, fold, horizon, and epoch identifiers so that observations from different evaluation conditions remain distinguishable.

Experiment identity includes the recipe, data content and splits, training settings, seed, precision, backend, and runtime source. Descriptive names and hypothesis wording do not change that identity. Cached measurements require matching recipes, data, settings, and verified artifacts; they neither provide independent evidence nor advance the no-improvement counter. Recovery verifies inputs, artifacts, and execution ownership without resetting budgets. Saved models contain weights, recipes, preprocessing, inference code, and evaluation metadata for prediction independently of the LLM.

\subsection{Core Research Prompts}
\label{app:core-prompts}
The following instructions summarize the main responsibilities in the research cycle. Task evidence, retrieved experience, component schemas, and available tools are supplied at runtime. Planning and design support the outer loop; optimization supports the inner loop, and reflection connects their experimental outcomes.

\phantomsection
\label{prompt:hypo-planner}
\textbf{Hypothesis Planning.} Initially propose 2--3 testable hypotheses from diagnostics, literature, and architecture-level experience. Record mechanisms, expected effects, supporting evidence, applicability assumptions, and falsification criteria. Use the supplied reference in Alpha and HMC; propose an initial design without a preset reference in Zero. Use experimental history, uncertainty, and the remaining budget to revise or select a hypothesis. Retain untested alternatives as unresolved.

\phantomsection
\label{prompt:model-designer}
\textbf{Model Design.} Translate the selected hypothesis into $p_i$ and $a_i$ using compatible components. Specify input/output requirements, representations, branches, and the initial training configuration in a complete recipe. For controlled comparisons, preserve undeclared settings from the named reference, which need not be the current best model. Correct implementation errors without changing the scientific intervention; return incompatible designs for outer-loop revision before further training attempts.

\phantomsection
\label{prompt:training-optimizer}
\textbf{Training Optimization.} Hold $d_i$ and its hypothesis fixed. Use optimization-level experience together with validation results, convergence curves, and error analysis to adjust $o_{i,j}$ and $c_{i,j}$. Record every attempt and select $\hat{f}_i$ from successfully trained and validated models by validation MSE. Return evidence to the outer loop when progress requires changing preprocessing or architecture.

\phantomsection
\label{prompt:reflection}
\textbf{Experiment Reflection.} Compare executed recipes, validation metrics, and curves with the declared reference. Check implementation, evaluation consistency, training adequacy, and confounding changes before assessing a mechanism. Distinguish execution failure from a completed evaluation with poor forecasts. Record training diagnoses for the inner loop and architectural assessments for the outer loop. Save conditional lessons with run evidence, retaining negative, conflicting, and unresolved outcomes.

\begin{table*}[t]
    \centering
    \caption{Dataset statistics and forecasting settings. $L$: input length; $H$: forecast length; $C$: input variables; $N_{\mathrm{test}}$: test samples. Each task has one target ($C_y=1$).}
    \label{tab:appendix-dataset-statistics}
    \begin{tabular*}{\textwidth}{@{\extracolsep{\fill}}llrrrrrrr@{}}
        \toprule
        Dataset & Frequency & Train & Validation & Test & $L$ & $H$ & $C$ & $N_{\mathrm{test}}$ \\
        \midrule
        ETTh1 & Hourly & 10,452 & 3,484 & 3,484 & 96 & 96 & 7 & 3,389 \\
        GEF-L & Hourly & 33,600 & 4,800 & 9,600 & 96 & 96 & 26 & 9,505 \\
        Windy & 15 min & 16,867 & 2,410 & 4,819 & 96 & 96 & 7 & 4,724 \\
        Sunny & 15 min & 16,867 & 2,410 & 4,819 & 96 & 96 & 7 & 4,724 \\
        MOPEX & Daily & 8,568 & 1,224 & 2,448 & 96 & 96 & 5 & 2,353 \\
        BE & Hourly & 10,147 & 1,450 & 2,899 & 168 & 24 & 3 & 2,876 \\
        DE & Hourly & 10,147 & 1,450 & 2,899 & 168 & 24 & 3 & 2,876 \\
        FR & Hourly & 10,147 & 1,450 & 2,899 & 168 & 24 & 3 & 2,876 \\
        NP & Hourly & 10,147 & 1,450 & 2,899 & 168 & 24 & 3 & 2,876 \\
        PJM & Hourly & 10,147 & 1,450 & 2,899 & 168 & 24 & 3 & 2,876 \\
        \bottomrule
    \end{tabular*}
\end{table*}

\begin{table}[!t]
    \centering
    \caption{Reasoning backbones. The best researcher depends on the forecasting task.}
    \label{tab:backbones}
    \setlength{\tabcolsep}{1.5pt}
    \renewcommand{\arraystretch}{0.85}
    \resizebox{\linewidth}{!}{%
    \begin{tabular}{@{}lrrrrrrrr@{}}
    \toprule
    Method & \multicolumn{2}{c}{ETTh1} & \multicolumn{2}{c}{MOPEX} & \multicolumn{2}{c}{NP} & \multicolumn{2}{c}{PJM} \\
    \cmidrule(lr){2-3} \cmidrule(lr){4-5} \cmidrule(lr){6-7} \cmidrule(lr){8-9}
     & MSE & MAE & MSE & MAE & MSE & MAE & MSE & MAE \\
    \midrule
    DeepSeek-V4.1Flash & \textbf{0.1004} & \textbf{0.2434} & \textbf{0.9812} & \textbf{0.5443} & \textbf{0.3546} & \textbf{0.3618} & \underline{0.1849} & \underline{0.3093} \\
    GPT-5.6 & \underline{0.1008} & \underline{0.2436} & \underline{1.0114} & 0.5649 & \underline{0.3635} & \underline{0.3717} & \textbf{0.1810} & \textbf{0.3056} \\
    GLM-5.2 & 0.1025 & 0.2473 & 1.0274 & \underline{0.5634} & 0.3821 & 0.3766 & 0.1899 & 0.3125 \\
    LongCat-2.0 & 0.1011 & 0.2440 & 1.0387 & 0.5691 & 0.4193 & 0.4010 & 0.1866 & 0.3110 \\
    \bottomrule
    \end{tabular}%
    }
\end{table}

\section{Hierarchical Experience Memory and Reuse}
\label{app:research-accumulation}

\subsection{Resources, Consolidation, and Retrieval}
Architecture-level experience retains mechanisms, prior hypotheses, design interventions, outcomes, and applicability conditions for outer-loop discovery. Optimization-level experience retains training configurations, convergence patterns, diagnoses, and tuning outcomes for the inner loop. Successful strategies guide subsequent choices, while unsuccessful paths discourage repeating ineffective adjustments under similar conditions. Both levels support decisions within and across tasks.

Knowledge, experience, and modeling assets are complementary resources supporting this hierarchy. Knowledge cards preserve immutable source records and applicability conditions; feedback links particular card revisions to experiments without rewriting the original findings. Experience links conditional lessons to factual records of interventions, verification, metrics, failures, and artifacts. Each completed task contributes its best recipe and architecture, selected by validation MSE, with source-task evidence. Trained weights remain in the source forecasting package, and the LLM's parameters remain unchanged.

Consolidation follows $M'=G(M,S)$ in Equation~\ref{eq:resource-consolidation}. Explicit writes attach run evidence and applicability conditions; narrative summaries alone do not create entries. Retrieval spans task, project, and shared scopes, combining intent, component, and condition matching with BM25-style scoring. Compact results expose lessons and evidence status, with underlying records available for inspection. Contradicted or unknown conditions preclude direct adoption. New experiments may narrow a finding's scope or leave conflicting observations unresolved. Failures and plateaus support diagnosis without directly refuting an architecture; mechanistic attribution requires controlled comparisons.

For design reuse, the outer loop adapts a retrieved recipe to the target interface, preprocessing, and architecture. After compatibility checks, the inner loop trains with newly initialized weights and optimizes training settings. Target-task validation guides selection and revision; source designs, adaptations, and outcomes remain linked in $S$.

\subsection{Memory Initialization for Evaluation}
\label{app:memory-evaluation}
Main results and component ablations use target-specific resources from cold-start research on the other nine datasets, subject to each variant's access restrictions. Source-memory construction permits literature retrieval and starts with empty experience and modeling-asset stores. Target evaluations do not form a transfer sequence or contribute to their own source-memory pools.

Figure~\ref{fig:sequential-research-accumulation} separately evaluates accumulation. Checkpoint 0 has no source-task memory; checkpoints 1--4 cumulatively add GEF-L, Windy, BE, and FR. ETTh1, MOPEX, NP, and PJM are the targets. Variants share source histories and a 25-round target budget per checkpoint. No Memory disables cross-task reuse. Raw Memory retains extracted knowledge, experience, and designs without formalization, using Full Memory's retrieval procedure. Full Memory organizes resources with evidence and applicability conditions. Checkpoints count source datasets, not research rounds.

\section{Datasets and Experimental Protocols}
\label{app:experimental-protocols}

\subsection{Datasets and Information Access}
\label{app:datasets}
EPF's BE, DE, FR, NP, and PJM~\cite{lago2021epf} pair hourly prices with load, generation, or wind-power forecasts. ETTh1~\cite{zhou2021informer} pairs transformer oil temperature with load measurements; GEF-L~\cite{hong2016gefcom2014} pairs demand with 25 weather-station temperatures. Windy and Sunny from the iFLYTEK AI Challenge provide wind/solar power and meteorological variables. MOPEX~\cite{schaake2006mopex} pairs streamflow with precipitation, evaporation, and temperature. All covariates are restricted to observed-past inputs; known-future and static inputs are unavailable for these forecasting tasks.

Table~\ref{tab:appendix-dataset-statistics} reports observations and forecasting settings. Training, validation, and test ratios are $6{:}2{:}2$ for ETTh1 and $7{:}1{:}2$ otherwise. A 96-step horizon spans one day for Windy/Sunny, four days for ETTh1/GEF-L, and 96 days for MOPEX.

\subsection{Training and Evaluation}
\label{app:baselines}
\label{app:evaluation}
Section~\ref{sec:experimental-setup} specifies the LLM settings, device, baselines, and evaluation criteria. Each task uses the single horizon in Table~\ref{tab:appendix-dataset-statistics}. Supported optimizers are Adam, AdamW, and SGD; point losses are MSE, MAE, and Huber, with matching losses required for distributional heads. Recipes resolve defaults, gradient clipping, scheduling, and epoch, patience, or time limits.

Checkpoint selection, training-level early stopping, and candidate selection use validation point-forecast MSE regardless of the training loss. The runner restores the best available checkpoint and rejects runs with no optimization update. Training stopping is separate from research stopping. Mamba runs record requested and actual scan backends. Held-out MSE and MAE assess the selected model; test observations are excluded from research decisions and from all validation folds.

\subsection{MOPEX Validation}
\label{app:analysis-settings}
MOPEX selection uses four-fold expanding-window validation with seed 42 before held-out testing. The runner reports the mean and population standard deviation of fold-level average MSE; variation reflects temporal blocks, not independent seeds. A separate fit on the original training/validation split produces the retained checkpoint.

\subsection{Ablation Protocols}
\label{app:ablation-protocols}
In Table~\ref{tab:evidence-ablation}, all variants use Alpha initialization from a human-designed reference. The w/o outer loop variant takes the first newly generated candidate, excluding the reference model, and fixes its preprocessing and architecture during inner-loop training optimization. The w/o inner loop variant retains architecture discovery but evaluates each candidate once under a fixed loss and training configuration, following the task's validation protocol.

In Table~\ref{tab:memory-ablation}, all variants retain both loops and current feedback. The w/o Architecture-level and w/o Optimization-level variants block historical experience for hypothesis planning and design revision, or training optimization, respectively; w/o Memory blocks both levels of historical experience.

In Table~\ref{tab:knowledge-ablation}, w/o Extraction uses original literature text for hypothesis planning: new retrieval bypasses extraction, and retained knowledge is accessed through its original sources. The w/o Knowledge variant disables new literature retrieval and retained literature knowledge while preserving experience and model assets.

For Figure~\ref{fig:cross-dataset-evolution}, the knowledge variant blocks stored cards but permits new retrieval; the experience variant excludes cross-task experience; and the modeling-asset variant disables retained design reuse. New literature and within-task experience remain available. These restrictions differ from the cumulative source exposure examined across checkpoints in Figure~\ref{fig:sequential-research-accumulation}.

\section{Reasoning Backbone Analysis}
\label{app:reasoning-backbones}
\label{sec:research-configuration-analysis}

Table~\ref{tab:backbones} compares the forecasting models returned by four LLM researchers. DeepSeek-V4.1Flash achieves the lowest MSE and MAE on ETTh1, MOPEX, and NP; GPT-5.6 leads on PJM. Secondary rankings can differ by metric: on MOPEX, GPT-5.6 has lower MSE than GLM-5.2, whereas GLM-5.2 has lower MAE. The LLM guides evidence interpretation, hypothesis formulation, design revisions, and validation-based selection. This comparison measures the forecasting outcomes of that research process. Once selected, each forecasting model performs numerical prediction independently of the LLM during inference.

\end{document}